%% file: main.tex
\documentclass{article}

\PassOptionsToPackage{numbers, sort&compress}{natbib}

\usepackage[preprint]{neurips_2022}

\usepackage[utf8]{inputenc} 
\usepackage[T1]{fontenc}    
\usepackage{hyperref}       
\usepackage{url}            
\usepackage{booktabs}       
\usepackage{amsfonts}       
\usepackage{nicefrac}       
\usepackage{microtype}      
\usepackage{xcolor}         

\usepackage[utf8]{inputenc}
\usepackage{amsmath,amsfonts,amssymb,amsthm,bbm}
\usepackage{dsfont}
\usepackage{booktabs} 

\usepackage{bm}
\usepackage{graphicx}
\usepackage{wrapfig}
\include{mysymbols}
\usepackage{algorithm}
\usepackage{algorithmic}
\usepackage{caption}
\usepackage{diagbox}
\usepackage{multirow}
\usepackage{lipsum}
\usepackage{pifont}
\usepackage{subcaption}

\usepackage{silence}
\hypersetup{colorlinks,
citecolor=blue,
}

\def\beingshort
\usepackage[titletoc]{appendix}

\title{Mutual Information Divergence: A Unified Metric for Multimodal Generative Models}

\author{%
  Jin-Hwa Kim\thanks{Corresponding author.} \hspace{1em}
  Yunji Kim \hspace{1em}
  Jiyoung Lee\\
  NAVER AI Lab\\
  Republic of Korea \\
  \texttt{\{j1nhwa.kim,yunji.kim,lee.j\}@navercorp.com} \\
  \AND
  Kang Min Yoo\\
  NAVER AI Lab \& CLOVA, SNU AIIS\\
  Republic of Korea\\
  \texttt{kangmin.yoo@navercorp.com}
  \And
  Sang-Woo Lee\\
  NAVER AI Lab \& CLOVA, KAIST AI\\
  Republic of Korea\\
  \texttt{sang.woo.lee@navercorp.com}
}

\begin{document}

\maketitle

\begin{abstract}
Text-to-image generation and image captioning are recently emerged as a new experimental paradigm to assess machine intelligence. They predict continuous quantity accompanied by their sampling techniques in the generation, making evaluation complicated and intractable to get marginal distributions.
Based on a recent trend that multimodal generative evaluations exploit a vison-and-language pre-trained model, we propose the negative Gaussian cross-mutual information using the CLIP features as a unified metric, coined by Mutual Information Divergence (MID). 
To validate, we extensively compare it with competing metrics using carefully-generated or human-annotated judgments in text-to-image generation and image captioning tasks. 
The proposed MID significantly outperforms the competitive methods by having consistency across benchmarks, sample parsimony, and robustness toward the exploited CLIP model. 
We look forward to seeing the underrepresented implications of the Gaussian cross-mutual information in multimodal representation learning and the future works based on this novel proposition. 
\end{abstract}

\section{Introduction}
\input{1_intro}

\section{Related work}
\input{2_related_work}

\section{Method}
\label{sec:Method}
\input{3_method}

\section{Experiment}
\label{sec:experiment}
\input{4_experiment}


\section{Conclusion}
\input{6_conclusions}

\begin{ack}
We sincerely thank Dongyoon Han for reviewing our manuscript and providing helpful comments. Also, we give thanks to Jung-Woo Ha for early discussions and suggestions throughout the project.
The NAVER Smart Machine Learning (NSML) platform~\cite{NSML} has been used in the experiments.
\end{ack}

\bibliography{mulgen}
\bibliographystyle{unsrtnat}

\if@preprint
\section*{Checklist}

\begin{enumerate}

\item For all authors...
\begin{enumerate}
  \item Do the main claims made in the abstract and introduction accurately reflect the paper's contributions and scope?
    \answerYes{}
  \item Did you describe the limitations of your work?
    \answerYes{Possible numerical issues in \sect~\ref{sec:Method}.}
  \item Did you discuss any potential negative societal impacts of your work?
    \answerNA{}
  \item Have you read the ethics review guidelines and ensured that your paper conforms to them?
    \answerYes{}
\end{enumerate}

\item If you are including theoretical results...
\begin{enumerate}
  \item Did you state the full set of assumptions of all theoretical results?
    \answerYes{}
        \item Did you include complete proofs of all theoretical results?
    \answerYes{\Appendix~\ref{appendix:proofs}}
\end{enumerate}

\item If you ran experiments...
\begin{enumerate}
  \item Did you include the code, data, and instructions needed to reproduce the main experimental results (either in the supplemental material or as a URL)?
    \answerYes{}
  \item Did you specify all the training details (e.g., data splits, hyperparameters, how they were chosen)?
    \answerYes{}
        \item Did you report error bars (e.g., with respect to the random seed after running experiments multiple times)?
    \answerYes{}
        \item Did you include the total amount of compute and the type of resources used (e.g., type of GPUs, internal cluster, or cloud provider)?
    \answerNA{}
\end{enumerate}

\item If you are using existing assets (e.g., code, data, models) or curating/releasing new assets...
\begin{enumerate}
  \item If your work uses existing assets, did you cite the creators?
    \answerYes{}
  \item Did you mention the license of the assets?
    \answerYes{See the reference. All assets are available to public.}
  \item Did you include any new assets either in the supplemental material or as a URL?
    \answerNA{}
  \item Did you discuss whether and how consent was obtained from people whose data you're using/curating?
    \answerNA{}
  \item Did you discuss whether the data you are using/curating contains personally identifiable information or offensive content?
    \answerNA{}
\end{enumerate}

\item If you used crowdsourcing or conducted research with human subjects...
\begin{enumerate}
  \item Did you include the full text of instructions given to participants and screenshots, if applicable?
    \answerYes{\Appendix~\ref{appendix:amt}}
  \item Did you describe any potential participant risks, with links to Institutional Review Board (IRB) approvals, if applicable?
    \answerNA{}
  \item Did you include the estimated hourly wage paid to participants and the total amount spent on participant compensation?
    \answerYes{\Appendix~\ref{appendix:amt}}
\end{enumerate}

\end{enumerate}
\else
\fi

\newpage
\setcounter{section}{0}
\renewcommand\thesection{\Alph{section}}
\renewcommand\thesubsection{\thesection.\arabic{subsection}}

\input{appendix}

\end{document}

%% file: mysymbols.tex
\newcommand{\pmi}{MID~}
\newcommand{\cond}{\vx}
\newcommand{\fake}{\vy}
\newcommand{\x}{\vx}
\newcommand{\y}{\vy}
\newcommand{\z}{\vz}
\newcommand{\fcond}{f_\vx}
\newcommand{\ffake}{f_\vy}

\newcommand{\T}{\intercal}
\newcommand{\E}{\mathbb{E}}
\newcommand{\R}{\mathbb{R}}
\newcommand{\N}{\mathcal{N}}

\newcommand{\tr}{\mathrm{tr}}

\newcommand{\I}{\mathbb{I}}
\newcommand{\One}{\mathds{1}}

\newcommand{\vx}{\mathbf{x}}
\newcommand{\vy}{\mathbf{y}}
\newcommand{\vz}{\mathbf{z}}
\newcommand{\vr}{\mathbf{r}}
\newcommand{\mX}{\mathbf{X}}
\newcommand{\mY}{\mathbf{Y}}

\newcommand{\mR}{\mathbf{R}}

\newcommand{\eg}{\textit{e.g.}}

\newcommand{\etc}{\textit{etc}}

\newcommand{\eqn}{Equation}

\newcommand{\fig}{Figure}
\newcommand{\tbl}{Table}
\newcommand{\sect}{Section}
\newcommand{\Appendix}{Appendix}

\newcommand{\U}{\underline}

\newcommand{\C}[1]{\scriptsize$\pm${#1}}
\newcommand{\X}[1]{\scriptsize{\color{white}$\pm$#1}}
\newcommand{\nz}{{\color{white}0}}  
\newcommand{\mz}{{\color{white}-0}}  
\newcommand{\mn}{{\color{white}-}}  

\newcommand{\paragrapht}[1]{\noindent\textbf{#1}}  

\newcommand{\cmark}{\ding{51}}
\newcommand{\xmark}{\ding{55}}

%% file: 1_intro.tex
A multimodal generative model, including text-to-image generation~\cite{frolov2021mulgen} and image captioning~\cite{chen2015coco} models, is an emerging research topic showing interpretative multimodal understanding, text or image retrieval, machine creativity, \etc.
The gist of learning multimodal generative models is to understand how to connect one modality to the other and generate the corresponding representations following the desired data distribution.
However, measuring the distance or divergence between the model and data distributions is generally intractable due to the finite data and generation cost. Therefore, the proposed metrics attempt to approximate it with polynomial-sized data and generated samples~\cite{thanh2020toward}.

For text-to-image generation, the widely-used metrics are Inception Score (IS)~\cite{salimans2016is} and Fréchet Inception Distance (FID)~\cite{Heusel2017}. These metrics are originally proposed for non-conditional generative models, which are repurposed to measure the distance between the data and model conditional distributions using a validation split. This idea was supported by the effectiveness of deep features as a perceptual metric~\cite{zhang2018unreasonable}, although these metrics use the Inception V3~\cite{szegedy2016inceptionv3}. Not surprisingly, there are the attempts to develop the more robust metrics using multimodal pre-trained models, such as object detectors~\cite{hinz2020soa}, image captioning models~\cite{hong2018inferring}, and vision-and-language pre-trained models~\cite{Radford2021,park2021benchmark}.
Significantly, the image captioning method transforms text-to-image measurement into image-to-text measurement, providing a different viewpoint with cyclic consistency.

For image-to-text generation, or image captioning, the COCO Caption Evaluation toolkit seems to be the standard to measure the divergence from ground-truth captions having BLEU~\cite{papineni2002bleu}, METEOR~\cite{banerjee2005meteor}, ROUGE~\cite{lin2004rouge}, CIDEr~\cite{vedantam2015cider}, and SPICE~\cite{anderson2016spice}. Similar to the IS and FID of text-to-image generation metrics, these metrics are merely n-gram-based statistical methods neglecting conditional images.
The breakthrough in improving the correlation with human judgment comes from the utilization of the pre-trained vision-and-language models, \eg, TIGEr~\cite{jiang2019tiger}, ViLBERTScore-F~\cite{lee2020vilbertscore}, and RefCLIP-S~\cite{Hessel2021}.
We speculate that these two directional metrics are getting closer to measuring the generative divergence of text-image alignment.
 
This paper proposes a unified metric for multimodal generative models.
In probability theory and information theory, mutual information (MI) measures how much one random variable tells us about the other.
In multimodal generation, the MI of two modalities quantitatively measures how much the generated is well-aligned with the condition.
From this motive, we propose to use the Gaussian mutual information where the probability distributions are defined by the means and covariances of visual and textual features and borrow the idea of cross-mutual information~\cite{bugliarello2020xmi} to measure the MI divergence from the real data distribution, which is the expectation of point-wise mutual information with respect to evaluating samples.
Surprisingly, the proposed method outperforms previous works with significant margins on the assorted benchmarks of text-to-image generation and image captioning, including standard human judgment correlation benchmarks.

\sect~\ref{sec:related_work} introduces the previous works on text-to-image generation and image captioning metrics, and a previous work on cross-mutual information. \sect~\ref{sec:Method} describes the proposed method defining the continuous mutual information using multivariate Gaussian distributions and the negative cross-mutual information, which is the Mutual Information Divergence (MID) what we term.
\sect~\ref{sec:experiment} consists of two parts, evaluation on text-to-image generation and image captioning evaluation, including related discussions.
\sect~\ref{sec:conclusions} concludes the work with remarks.

We summarize our contributions as follows:
\begin{itemize}
  \item To the best of our knowledge, we firstly propose the negative cross-mutual information under the Gaussian assumption as a unified metric for multimodal generative models.
  \item We provide three theoretical analyses on the proposed method MID, out-of-distribution detecting by the squared Mahalanobis distances, bias and variance decomposition, and its relation to the Kullback-Leibler divergence.
  \item We achieve the state-of-the-art on text-to-image generation and image captioning benchmarks including the generated and human Likert-scale judgment correlations, visual reasoning accuracy, Flickr8K-Expert, Flickr8K-CF, Pascal-50S, and FOIL hallucination detection.
\end{itemize}

\begin{figure}[t!]
  \begin{center}
    \includegraphics[width=\textwidth]{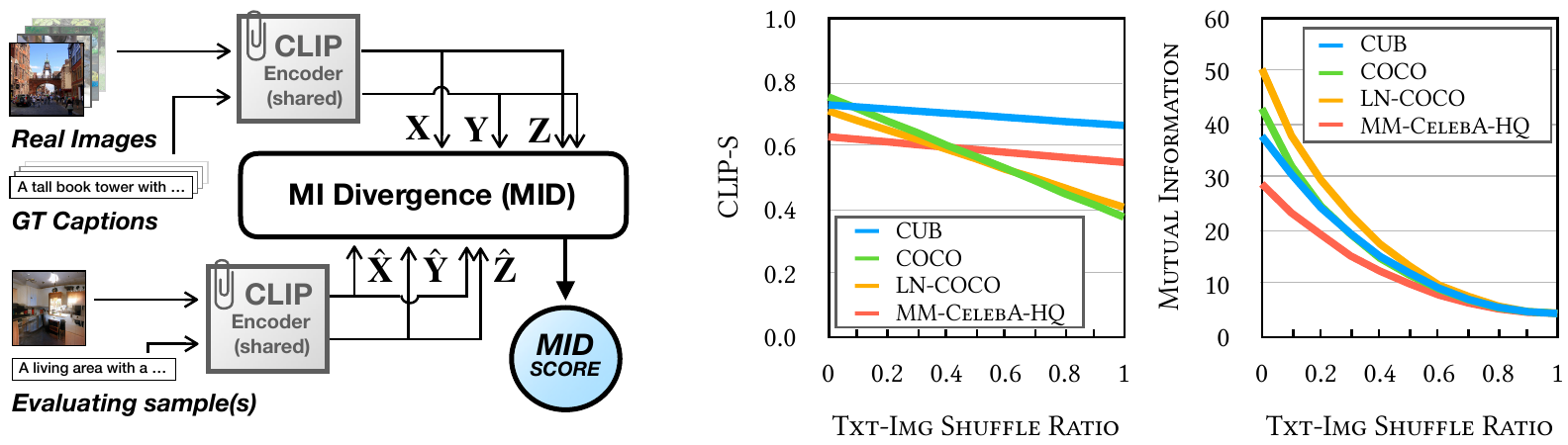}
  \end{center}
  \caption{\textbf{Left.} The schematic diagram of our proposed method. \textbf{Right.} The consistent property of Gaussian MI is in stark contrast to CLIP-S~\cite{Hessel2021}.
We vary the ratio of text-image shuffling where the counterparts in the selected pairs are deliberately shuffled, depicting misalignment. For CLIP-S, we observe varying slopes across the datasets depending on data domain while MI shows a relatively consistent tendency. The CUB and MM-CelebA-HQ describe birds and human faces, respectively, having narrow domains compared to COCO and LN-COCO with various objects.}
  \label{fig:one}
\end{figure}

%% file: 2_related_work.tex
\label{sec:related_work}
\subsection{Metrics for assessing text-to-image generation}

\paragrapht{Traditional metrics.} 
One of the widely used metrics is the Fréchet Inception Distance (FID)~\cite{Heusel2017} that measures the distributional difference between synthetic (fake) and real-world images (real). 
Training and validation distributions are independent, making a model that fails to match the conditional distributions if it does not reflect the given textual information. 
Although it measures fidelity along with Inception Score~\cite{salimans2016is}, 
it cannot directly measure the alignment of text and image.
The alternative metrics~\cite{sajjadi2018pr,kynkaanniemi2019pr,naeem2020reliable} are proposed to evaluate the fidelity and diversity. 
\paragrapht{Text-to-image metrics.} 
Dedicated to assessing text-to-image generation, the R-Precision exploits the Deep Attentional Multimodal Similarity Model (DAMSM)~\cite{xu2018attngan} to calculate the top-1 retrieval accuracy from one hundred text candidates for the generated image as a query. Besides, the CLIP R-Precision~\cite{park2021benchmark} exploits the CLIP~\cite{Radford2021} showing a better retrieval performance and human judgment correlation. However, false-negative candidates (accidentally correlated) or strong negative candidates (totally unrelated) may interfere with the accurate assessment~\cite{chun2022eccv}. To evaluate the quality of individual objects, the SOA~\cite{hinz2020soa} attempts to measure the object detection accuracy using the YOLOv3~\cite{redmon2018yolov3}
based on the object classes that appeared in the text but cannot consider other factors.
Caption generation~\cite{hong2018inferring} is another approach using the vision-and-language pre-trained model. The motivation is the cyclic consistency that the generated caption from the generated image should match with the text for image generation. However, the model bias including object hallucination~\cite{rohrbach2018hallucination} and the accumulated errors from metric are drawbacks.
\paragrapht{Diagnostic datasets.} 
\citet{park2021benchmark} provide the curated splits of the CUB~\cite{wah2011cub,reed2016cubstr} and Flowers~\cite{nilsback2008flowers} to assess unseen color and shape compositions in the narrow domains.
DALL-Eval~\cite{cho2022dall} proposed a diagnostic dataset \textit{PaintSkills} to evaluate visual reasoning skills to assess models based on this dataset.
Since the dataset is generated from a 3D simulator using limited configurations, the data distributions deviate from other real-world datasets.

\subsection{Metrics for assessing image captioning}
\paragrapht{Reference-only metrics.} 
Borrowing from machine translation literature, image captioning models are evaluated using reference-based textual metrics. Usually, five references are used to measure in these metrics. BLEU-4~\cite{papineni2002bleu}, ROUGE-L~\cite{lin2004rouge}, and METEOR~\cite{banerjee2005meteor} are n-gram precision or recall-based metrics, CIDEr~\cite{vedantam2015cider} uses tf-idf weighting and stemming, while SPICE~\cite{anderson2016spice} uses semantic parsing and scene graph analysis. Notably, BERT-S++~\cite{yi2020improving} considers inter-reference variance using the fine-tuned BERTScore~\cite{zhang2019bertscore} for image captioning. \paragrapht{Reference-with-image metrics.} 
The recently proposed metrics are considering the images used for generating captions. TIGEr~\cite{jiang2019tiger} uses a pre-trained SCAN~\cite{lee2018scan} while ViLBERTScore-F~\cite{lee2020vilbertscore} uses a pre-trained ViLBERT~\cite{lu2019vilbert} exploiting the vision-and-language alignments from large-scale data and multiple tasks. Similarly, CLIP-S and RefCLIP-S~\cite{Hessel2021} uses the pre-trained CLIP, a more powerful vision-and-language model, outperforming the previous methods.
\paragrapht{Implications.}
CLIP R-Precision for text-to-image generation and RefCLIP-S for image captioning share the same motivation exploiting the same vision-and-language pre-trained model. Here, we remark on the unifying metrics in two different tasks (\eg, CLIP-S), text-to-image generation and image captioning, and propose a new unified metric for the multimodal generative models based on the continuous mutual information considering the covariances of two modality groups.

\subsection{Cross-mutual information in machine translation}
\citet{bugliarello2020xmi} proposed the cross-mutual information (XMI) as a metric of machine translation exploiting a probabilistic view in neural machine translation models. XMI is an analogue of mutual information for cross-entropy defined as $\text{XMI}(S \rightarrow T) = H_{q_{\text{LM}}}(T) - H_{q_{\text{MT}}}(T|S)$. $H_{q_{\text{LM}}}(T)$ denotes the cross-entropy of the target sentence $T$ under a language model $q_{\text{LM}}$ and $H_{q_{\text{MT}}}(T|S)$ is the cross-conditional entropy under a cross-lingual model $q_{\text{MT}}$.
In practice, they exploit two model distributions $q_{\text{LM}}(\mathbf{t})$ and $q_{\text{MT}}(\mathbf{t}|\mathbf{s})$ to approximate the XMI as follows:
$$
    \text{XMI}(S \rightarrow T) \approx - \frac{1}{N}\sum_{i=1}^{N} \log 
    \frac{
        q_{\text{LM}}(\mathbf{t}^{(i)})
    }{
        q_{\text{MT}}(\mathbf{t}^{(i)}|\mathbf{s}^{(i)})       
    }
$$
where $N$ denotes the number of held-out evaluating samples. Since language models consider a finite size of vocabulary, the cross-entropy can be efficiently approximated using the target sentences; however, it is limited to readily apply to other generative models.


%% file: 3_method.tex
\subsection{Continuous mutual information}
We introduce a unified metric for conditional generative models not depending on the modalities of \textit{condition} and \textit{generation}. 
To measure the alignment of condition and generation, we first consider the continuous mutual information of the condition $\cond$ and generation $\fake$ as follows:
\begin{align}
    I(\mX; \mY) = \E_{p(\cond, \fake)}\log\frac{p(\cond, \fake)}{p(\cond) p(\fake)} 
\end{align}
where the probability and joint probability distributions are multivariate Gaussian, which is the maximum entropy distribution for the given mean $\mu$ and covariance $\Sigma$~\cite{dowson1973maximum}.
The first two moments are used for practical reason~\cite{Heusel2017}.
The multivariate Gaussian distribution is defined as:
\begin{align}
    p(\vx) = \frac{1}{\sqrt{(2\pi)^D \det(\Sigma)}} \exp \big[ -\frac{1}{2}(\vx - \mu)^\T \Sigma^{-1} (\vx - \mu) \big]
\end{align}
where $D$ is the dimension of $\vx$. The mutual information with the Gaussian distributions is reduced to:
\begin{align}
    I(\mX; \mY) = \frac{1}{2}\log\Big( \frac{\det(\Sigma_\x) \det(\Sigma_\y)}{\det(\Sigma_\z)} \Big)
\end{align}
where $\Sigma_\x$ and $\Sigma_\y \in \R^{D \times D}$ are the covariances of the condition and generation and $\Sigma_\z \in \R^{2D \times 2D}$ is the covariance matrix of the concatenation $\z$ of $\cond$ and $\fake$ representing the joint distribution. The proof can be found in \Appendix~\ref{appendix:mi-proof}. Note that we use $\log\det(\Sigma) = \sum_i \log \lambda_i$ for numerical stability, where $\lambda_i$ is the eigenvalue of $\Sigma$.

For high dimensional data, we consider two encoders $\fcond$ and $\ffake$ to get $\tilde{\cond}$ and $\tilde{\fake}$, respectively, maximizing the mutual information of the encoded representations, $I(\tilde{\mX}; \tilde{\mY})$. 
Specifically, the image and text encoders of the CLIP~\cite{Radford2021} are used since these are pre-trained on 400M image-text pairs using the InfoNCE loss, maximizing a lower bound on mutual information~\cite{oord2018infonce}.
Without loss of generality, we use $\x, \y$ instead of $\tilde{\x}, \tilde{\y}$ to denote the feature vectors to calculate the moments.


\subsection{Point-wise mutual information for pair-wise evaluation}

Based on the previous continuous mutual information, we derive the \textit{point-wise mutual information (PMI)} for pair-wise evaluation. Please see \Appendix~\ref{appendix:mi-proof} for the detail. The PMI is defined as:
\begin{align}
    \text{PMI}(\x; \y) = I(\mX; \mY) + \frac{1}{2} \big(
        D_M^2(\x) + D_M^2(\y) - D_M^2(\z)
    \big). \label{eqn:pmi}
\end{align}
where $D_M^2$ denotes the squared Mahalanobis distance (SMD), $D_M^2(\x) = (\x - \mu_\x)^\T \Sigma_\x^{-1} (\x - \mu_\x)$ where $\mu_\x$ and $\Sigma_\x$ are the mean and covariance of $\x$, and similarly for $D_M^2(\y)$ and $D_M^2(\z)$. $\vz$ denotes $[\x; \y]$.
The MI is from the normalization of the Gaussians and the SMDs are from the exponential of the Gaussians.
The second term measures the difference between the distances, $D_M^2(\x) + D_M^2(\y)$ and $D_M^2(\z)$, assessing the deviation from the MI.
Notice that the expectation of the second term with respect to the sample distribution is zero (see \Appendix~\ref{appendix:mi-proof}). 

\subsection{Mutual Information Divergence: the expectation of PMI w.r.t evaluating samples}

We propose to use the expectation of PMI with respect to the evaluating sample $(\hat{\x}, \hat{\y})$, measuring the divergence from the ground-truth or reference samples $(\mX, \mY)$. 
The metric is defined as follows:
\begin{align}
    \E_{(\hat{\x}, \hat{\y}) \sim \mathcal{D}} \text{PMI}(\hat{\x}; \hat{\y}) = I(\mX; \mY) + \frac{1}{2} \E_{(\hat{\x}, \hat{\y}) \sim \mathcal{D}} \big[
        D_M^2(\hat{\x}) + D_M^2(\hat{\y}) - D_M^2(\hat{\z})
    \big]. \label{eqn:exp_pmi}
\end{align}
where $\mathcal{D}$ and $(\hat{\x}, \hat{\y})$ denote the set of evaluating samples and a pair of evaluating sample, respectively, $\hat{\vz}$ denotes $[\hat{\x}; \hat{\y}]$. 
Notice that the expectation of $D_M^2(\hat{\x})$ can be decomposed to the bias and variance terms as follows (\Appendix~\ref{appendix:bias-variance decompositon} for the proof):
\begin{align}
\label{eqn:expectation_of_smd}
    \E_{\hat{\x}} \big[ D^2_M(\hat{\x}) \big] 
        &= (\mu_{\hat{\x}} - \mu_{\x})^\intercal \Sigma_\x^{-1}( \mu_{\hat{\x}} - \mu_{\x}) + \tr\big(\Sigma_\x^{-1}(\Sigma_{\hat{\x}} - \Sigma_{\x})\big) + D
\end{align}
considering the mean and covariance deviations from the reference, $\N(\mu_\x, \Sigma_\x)$.
$\E_{\hat{\y}} \big[ D^2_M(\hat{\y}) \big]$ is $D$ when $\hat{\y} = \y$ as a generative condition since the two moments are equal to the counterparts.


By the way, we can show that $\E_{(\hat{\x}, \hat{\y}) \sim \mathcal{D}} \text{PMI}(\hat{\x}; \hat{\y})$ is related to the Kullback–Leibler divergence as follows (The proof can be found in \Appendix~\ref{appendix:kl-divergence}):
\begin{align}
    \E_{(\hat{\x}, \hat{\y}) \sim \mathcal{D}} \text{PMI}(\hat{\x}; \hat{\y}) 
        = & I(\hat{\mX}; \hat{\mY}) + D_{\text{KL}}(p(\hat{\x}) \parallel p(\x) ) - D_{\text{KL}}(p(\hat{\z}) \parallel p(\z) )
        \label{eqn:mid_kl_div}
\end{align}
For simplicity, we denote our proposed method $\E_{(\hat{\x}, \hat{\y}) \sim \mathcal{D}} \text{PMI}(\hat{\x}; \hat{\y})$ as {\bf Mutual Information Divergence (MID)}, comparable of the FID.
In practice, we use \eqn~\ref{eqn:exp_pmi} using the double-precision CLIP features.
For point-wise evaluation, we use the $\text{PMI}(\hat{\x}, \hat{\y})$ without the expectation.

%% file: 4_experiment.tex

\subsection{Evaluation on text-to-image generation}

\paragrapht{Implementation details.}
Without an explicit mention, we use the CLIP (ViT-L/14) to extract image and text embedding vectors. Note that it is crucial to use double-precision for numerical stability.

\input{tbl_likert}
\input{tbl_human}

\paragrapht{Generated Likert-scale judgments.}
To carefully assess the text-image alignment, we consider the four-scale alignment using the real and fake images from the COCO dataset~\cite{chen2015coco}.
We regard the real images as a four-point set, the fake images generated by the ground-truth captions as a three-point set, the fake images generated by the foiled captions~\footnote{For the details, please refer to the object hallucination section in \sect~\ref{sec:object_hallucination} and \fig~\ref{fig:vis_foil} in \Appendix.}~\cite{shekhar2017foil} as a two-point set, and the randomly sampled (misaligned) fake images as one-point set. 
We assume that the fake images generated by the foiled captions should be inferior compared with the fake images generated by the ground-truth captions because the model cannot exploit the critical information to generate key objects.
The current state-of-the-art LAFITE~\cite{zhou2021lafite} pre-trained on the COCO~\footnote{\url{https://github.com/drboog/Lafite}} is used for our text-to-image generation model.
Notice that when we evaluate the metrics, the ground-truth captions are used to measure the text-image alignment of the generated images. 
We believe this generated benchmark can be a proxy to human judgments with a careful manipulation of the fake images using the foiled captions.

We randomly sample 30K captions from the FOIL dataset~\cite{shekhar2017foil} to build 120K judgments. We report the Kendall's $\tau$ coefficient~\cite{kendall1938tau} to measure the rank correlation, a variant of $\tau_c$ or $\tau_b$ accounting for ties. 
In \tbl~\ref{tbl:likert-lafite}, our method consistently outperforms competing methods.
InfoNCE denotes the negative InfoNCE loss~\cite{oord2018infonce} calculating the softmax function over the 30K captions. Remind that the InfoNCE maximizes a lower bound on mutual information~\cite{oord2018infonce}. Since this is estimated using a smaller batch size when optimizing, it shows a limited capability as a metric for the text-image alignment.
Please refer to \tbl~\ref{tbl:likert-vqdiffusion} and \fig~\ref{tbl:likert-vqdiffusion} in \Appendix~for the VQ-Diffusion~\cite{gu2021vqdiffusion} benchmark.

\begin{figure}[t!]
  \begin{center}
    \includegraphics[width=\textwidth]{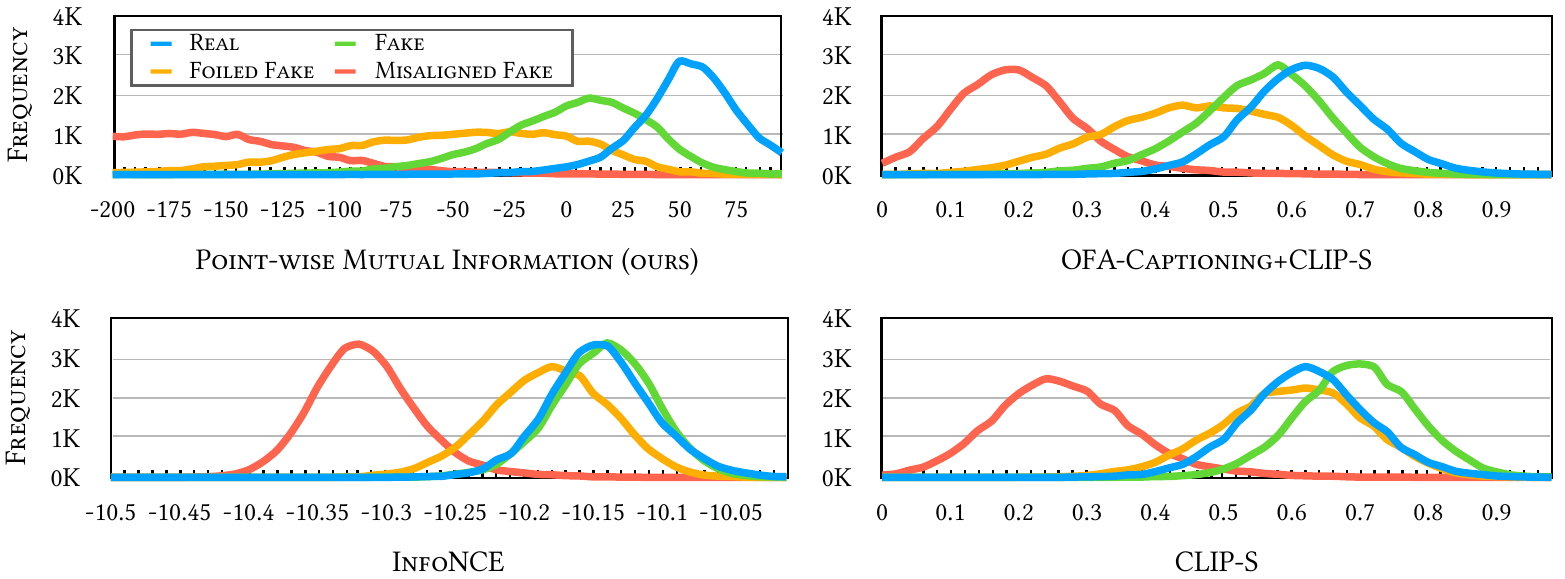}
  \end{center}
  \caption{The histogram shows the frequencies of the competing methods using the CLIP of ViT-L/14 for the four-scale judgments from the COCO dataset. 
For the detail of method, please see the text.
  }
  \label{fig:likert}
\end{figure}

\fig~\ref{fig:likert} (top left) shows the histogram of the frequencies of PMI showing the four categories. {\bf Fake} samples have generally lower values than {\bf Real}'s. {\bf Foiled Fake} has lower values than {\bf Real} and {\bf Fake} having a long tail. We observe that foiled caption broadly impacts the text-image alignment as we expected.
The negative PMI is often observed for the fake images, which are deviated from the distribution of real images. 
Whereas, the histogram of CLIP-S~\footnote{The CLIP-S~\cite{Hessel2021} is defined as $2.5 \times \cos(\x, \y)$ which is the scaled cosine similarity of the CLIP features.} (bottom right) and InfoNCE (bottom left) show {\bf Fake} samples where its scores are higher than {\bf Real}'s, while the overlapping areas of {\bf Fake} and {\bf Foiled Fake} (62.2\% and 61.6\% for CLIP-S and InfoNCE, respectively) are greater than PMI's (55.2\%), making it difficult to differentiate the degree of text-image alignment.
For the caption generation method~\cite{hong2018inferring}, we exploit the current state-of-the-art image captioning model of the OFA-Large~\cite{wang2022ofa} pre-trained on a huge mixture of publicly available datasets to generate captions for the generated images. Then, CLIP-S~\cite{Hessel2021} is used to assess the quality of image captioning. 
Notice that CLIP-S outperforms traditional image captioning metrics as shown in \sect~\ref{sec:evaluation_on_image_captioning}.


\paragrapht{Visual reasoning accuracy using the foiled caption trick.}
\input{tbl_reasoning}
Inspired by the foiled caption trick, we extend to four visual reasoning skills for object, count, color, and spatial relationship. For each category, we define a set of tokens, and we foiled those tokens in the caption by randomly swapping to the other token. We build three sets of images, the real images, the fake images, and the foiled fake images. We measure the accuracy that getting one point for the foiled fake images having the lowest score, or zero for the other cases.
\tbl~\ref{tbl:reasoning} shows that our method achieved the best performance across all categories. The runner-up was InfoNCE, while SOA was ineffective to differentiate among the two fake images and the real image.
Although DALL-Eval~\cite{cho2022dall} proposed to use a detector and its dedicated heads for the count, color, and spatial relationship tasks, this was limited to the 3D-generated images with their near-perfect detection ability.
Remind that the accuracy of random guessing is 33.3\%, where MID requires a powerful feature extractor of the CLIP ViT-L/14 to get meaningful performances on the count, color, and spatial relationship tasks.
For the detail, please refer to the text in \Appendix~\ref{appendix:reasoning}.

\paragrapht{Human Likert-scale judgment.}
\label{sec:amt}
We collect 10K one-to-four Likert-scale human judgments for 2K fake images using the LAFITE and VQ-Diffusion from the Amazon Mechanical Turk (AMT) for the fine-grained comparison with the competing metrics. For each image, we collect five annotations from unique workers taking its median for a reliable correlation measurement. \tbl~\ref{tbl:likert-amt} shows the consistent results with our generated judgment correlation benchmark.
Since this benchmark aims for the fine-grained judgment among only fake images, overall scores are relatively lower than the generated benchmark. All correlation results are significant having the p-value < 0.001.
For the details of the collection procedure and data statistics, please refer to \Appendix~\ref{appendix:amt}, where
\fig~\ref{fig:amt_vis} shows the visualization of some examples comparing with human judgment scores.
We also report the comp-t2i benchmark~\cite{park2021benchmark} results for the compositional evaluation of the CUB and Flower datasets in aspects of color and shape in \Appendix~\ref{appendix:comp-t2i}.

\subsubsection{Discussions}

\paragrapht{Consistent metric across datasets.} A distinct property of our method is the consistency across datasets. As shown in \fig~\ref{fig:one}, the cosine similarity-based method, CLIP-S suffers the inconsistent results. For example, the CUB~\cite{wah2011cub,reed2016cubstr} and MM-CelebA-HQ~\cite{xia2021celeba} have narrow domains, birds and human faces, respectively, which is prone to get a similar score for all samples in the datasets by cosine similarity. To validate our hypothesis, we vary the ratio of text-image shuffling using the real datasets where the counterparts in the selected pairs are deliberately shuffled, depicting misalignment. For CLIP-S, we observe the inconsistency depending on datasets, while MI shows a consistent tendency. Notice that the expectation of PMI is reduced to MI for the real images (\Appendix~\ref{appendix:mi-proof}).

\begin{figure}[t]
  \begin{center}
    \includegraphics[width=.85\textwidth]{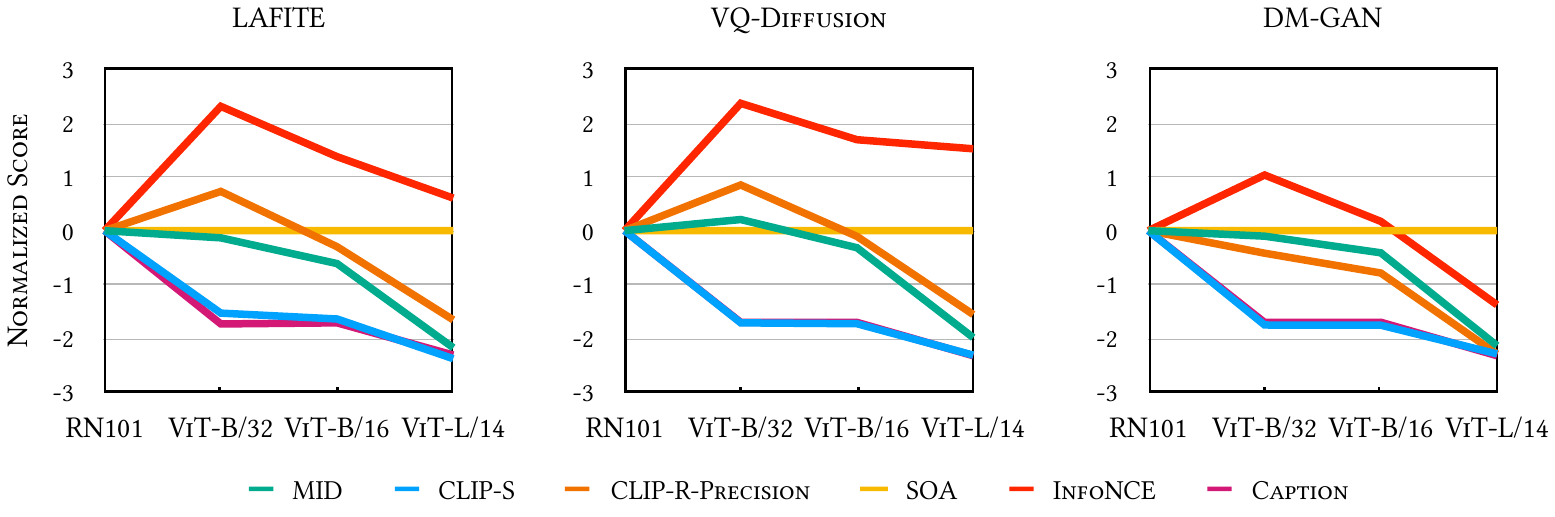}
  \end{center}
  \caption{The footprints of metrics across feature extractors using the normalized scores.
  }
  \label{fig:overfit}
\end{figure}

\paragrapht{Inspecting possible over-fitting with the CLIP features.}
\fig~\ref{fig:overfit} shows the normalized scores of metrics across feature extractors.
To get the normalized score, we subtract the score of RN101 and divide by the standard deviation of four scores across feature extractors reminiscent of z-score.
We expect the \textit{footprint} of metrics should be consistent across different generative models if the model is not over-fitted to the metrics.
LAFITE used both encoders of CLIP ViT-B/32 and VQ-Diffusion used the text encoder of CLIP ViT-B/32, while DM-GAN did none of them~\cite{zhu2019dmgan}.
InfoNCE and CLIP-R-Precision are related to the contrastive training losses (LAFITE, and DM-GAN for the DAMSM loss~\cite{xu2018attngan}), which may lead to drastic change of the normalized scoring signature across the feature extractors.
While the proposed MID was relatively stable across the generative models.

\paragrapht{Comparison with the state-of-the-art models using MID.}
\tbl~\ref{tbl:gans} in \Appendix~\ref{appendix:gans} exhibits the MID performance of the recent text-to-image generative models along with the other major metrics.

\subsection{Evaluation on image captioning}
\label{sec:evaluation_on_image_captioning}

\paragrapht{Implementation details.} 
For a fair comparison with the current state-of-the-art evaluation metric (RefCLIP-S), we use the same pre-trained CLIP (ViT-B/32) used in the prior work to extract image and caption embedding vectors. 
We use the images and the corresponding reference captions to build the covariance matrices $\Sigma_\vx, \Sigma_\vy$ and the joint covariance matrix $\Sigma_\vz$.
For the numerical stability of the inverse of covariance matrix, we replace $\Sigma_\vx^{-1}$ with $\tilde{\Sigma}_\vx^{-1} = (\Sigma_\vx + \epsilon \I)^{-1}$, which handles the near-zero eigenvalues of covariance. We found that $\epsilon$ of 5e-4 generally works across all benchmark evaluations, except for the FOIL benchmark where we used $\epsilon$ of 1e-15, which was slightly better.
Note that we use an identical prompt ``A photo depicts'' for all caption embeddings as employed in RefCLIP-S~\cite{Hessel2021}.

\input{tbl_flickr8k}

\paragrapht{Flickr8K-Expert and Flickr8k-CF.}
We measure the correlation of the proposed method with the Likert-scale judgments, which indicate the relative correctness of given captions. Fliker8K-Expert~\cite{hodosh2013f8expert} provides 17K human expert judgments for 5,664 images with a four-scale where the higher is better. Following prior works, we flatten all human judgments to a list of 16,992 (5,664$\times$3) samples, and we exclude 158 pairs where their captions appear in the reference set. Flickr8K-CF~\cite{hodosh2013f8expert} has 145K binary judgments from the CrowdFlower for 48K image-caption pairs. Each pair receives at least three judgments, and we take the proportion of positive as a corresponding score. Kendall's $\tau$ coefficient~\cite{kendall1938tau} measures the rank correlation, and $\tau_c$ and $\tau_b$ are used for Fliker8K-Expert and Flickr8K-CF, respectively. Although $\tau_c$ is more suitable when the underlying scales differ in two variables, we follow the previous works for a fair comparison.
{\tbl}s~\ref{tbl:flickr8k-expert} and \ref{tbl:flickr8k-cf} show the evaluation results. For both cases, our \pmi significantly improves the correlation of human judgments with 54.9 and 37.3, respectively.
Notably, \pmi improves further than RefCLIP-S, which uses the same vision-and-language pre-trained CLIP.

\begin{wrapfigure}{r}{0.3\textwidth}
  \small
  \begin{center}
    \vspace{-2.2em}
    \includegraphics[width=0.3\textwidth]{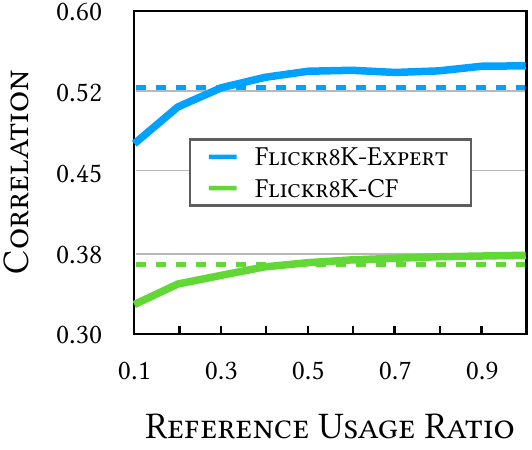}
  \end{center}
  \vspace{-2pt}
  \caption{Flickr8K parsimony.}
  \label{fig:parsimonious}
\end{wrapfigure}

\paragrapht{Flickr8K reference parsimony.}
Compared with the other methods, our method does not directly rely on the corresponding references, but through the mean and covariance. Therefore, we could exploit the sample statistics with a limited number of references. \fig~\ref{fig:parsimonious} shows the Kendall's $\tau$ correlation utilizing a subset of the available references. Interestingly, even though 30-40\% of images are available, it retains the majority of performance. The dashed lines indicate the correlations of RefCLIP-S, which exploits all references.
Notice that our method can be positioned between the reference-with-image and reference-free metrics.
Because a sufficient amount of samples are required to assess image captioning models, the sample statistics can be reliable from the sufficient samples.

\paragrapht{Pascal-50S.} For a different evaluation setting of accuracy, the Pascal-50S~\cite{vedantam2015cider} offers 4K pair-wise preference judgments between two captions, evenly splitting four categories, two human correct captions (HC), both human written, but one is incorrect (HI), one is from human, the other is by a model (HM), and both are generated by machine (MM). For each pair, there are 48 human judgments and the majority of votes decides which caption is preferred, where ties are broken randomly. As in the previous work~\cite{Hessel2021}, we randomly sample 5 references among 48 candidates and average over five evaluations. \tbl~\ref{tbl:pascal} shows the consistent results outperforming competitive methods. Except for HI, we achieve the state-of-the-art while the margin of HI is 0.1 having a near-perfect score of 99.7.

\begin{wrapfigure}{r}{0.33\textwidth}
  \small
  \begin{center}
    \vspace{-1.5em}
    \includegraphics[width=0.33\textwidth]{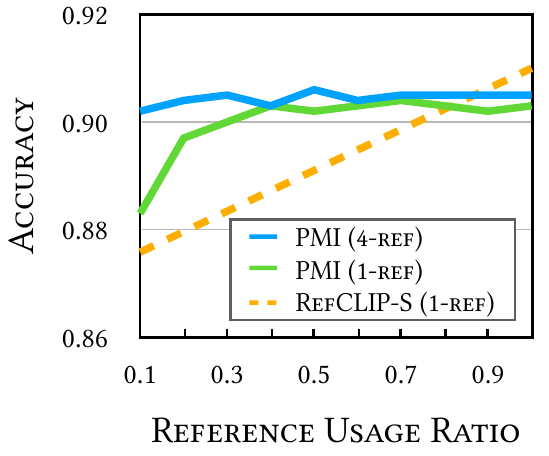}
  \end{center}
  \caption{FOIL parsimony.}
  \label{fig:foil}
\end{wrapfigure}

\input{tbl_pascal_foil}

\paragrapht{Object hallucination sensitivity.}
\label{sec:object_hallucination}
\citet{rohrbach2018hallucination} argue that image captioning models prone to generate the objects not presented in the image due to learned bias. To assess this aspect, the FOIL-COCO~\cite{shekhar2017foil} builds the carefully modified captions from the COCO captions~\cite{chen2015coco} by swapping a single noun-phrase, \eg, substituting ``cat'' for ``dog''. To measure the accuracy whether assigning a higher score to the ground-truth caption over the FOIL caption, we evaluate 32K test images with exclusive four reference captions of the COCO dataset. \tbl~\ref{tbl:foil} shows the competitive scores of our method. Since RefCLIP-S~\cite{Hessel2021} directly accesses the reference captions of the evaluating image, they can exploit the original words (before foiling) in the references, which is roughly 87\% in four references and 67\% in a randomly selected reference.
The RefCLIP-S is defined as follows:
\begin{align}
    \label{eqn:refclip-s}
    \text{RefCLIP-S}(\vx, \vy, \mR) = H\big(\text{CLIP-S}(\vx, \vy), \max(\max_{\vr \in \mR} \cos(\vr, \vy), 0)\big)
\end{align}
where $H$ is harmonic mean, $\vx$ and $\vy$ denotes image and caption embeddings, and $\mR$ is a set of the references.
\fig~\ref{fig:foil} supports this. Our method is robust via covariance estimation in both 1-ref (green line) and 4-ref (blue line), while RefCLIP-S is reducing to its reference-free version of CLIP-S degrading performance. We interpolate the scores of RefCLIP-S and CLIP-S as a reference (yellow line).
If we similarly utilize the term of $\max(\max_{\vr \in \mR} \cos(\vr, \vy), 0)$ as in \eqn~\ref{eqn:refclip-s}, \pmi gets 92.4 and 93.7, outperforming RefCLIP-S (\Appendix~\ref{appendix:extra-foil} for the detail, \fig~\ref{fig:vis_foil} for the visualization).

%% file: tbl_likert.tex
\begin{table*}[t!]
  \centering
  \small
  \caption{Generated Likert-scale judgment correlation using LAFITE. $\dagger$ uses the i.i.d. samples having at least one detected object, which was 88.3\% of samples, to calculate the SOA accuracy per image.} 
  \label{tbl:likert-lafite}
  \begin{tabular}{lccc}
  \toprule
  Method & Backbone & Kendall $\tau_c$ & Kendall $\tau_b$ \\ \midrule
  SOA$^\dagger$~\cite{hinz2020soa} & YOLO-V3 & 47.3 & 51.8 \\
  \midrule
  CLIP-S~\cite{Hessel2021} & CLIP (ViT-B/32) &  40.8 & 35.3 \\
  InfoNCE~\cite{oord2018infonce} & CLIP (ViT-B/32) & 44.1 & 38.2 \\
  CLIP-R-Precision~\cite{park2021benchmark} & CLIP (ViT-B/32) & 66.0 & 56.1 \\
  OFA-Captioning+CLIP-S~\cite{wang2022ofa,Hessel2021} & OFA-Large + CLIP (ViT-B/32) & 72.0 & 62.3 \\
  \midrule
  CLIP-S~\cite{Hessel2021} & CLIP (ViT-L/14) &  52.2 & 45.2 \\
  InfoNCE~\cite{oord2018infonce} & CLIP (ViT-L/14) & 64.8 & 56.1 \\
  CLIP-R-Precision~\cite{park2021benchmark} & CLIP (ViT-L/14) & 69.6 & 58.1 \\
  OFA-Captioning+CLIP-S~\cite{wang2022ofa,Hessel2021} & OFA-Large + CLIP (ViT-L/14) & 73.7 & 63.8 \\
  \midrule
  \pmi (ours) & CLIP (ViT-B/32) & \underline{74.6} & \underline{64.6} \\
  \pmi (ours) & CLIP (ViT-L/14) & \bf{87.3} & \bf{75.6} \\
  \bottomrule
  \end{tabular}
\end{table*}

%% file: tbl_human.tex
\begin{table*}[t!]
  \centering
  \small
  \caption{AMT human Likert-scale judgment correlation. Using the 2K fake images from LAFITE and VQ-Diffusion, we collected fine-grained text-image alignment judgments and measure the judgment correlation of competing metrics. $\dagger$ 90.2\% of samples having at least one detected object.} 
  \label{tbl:likert-amt}
  \begin{tabular}{lccc}
  \toprule
  Method & Backbone & Kendall $\tau_c$ & Kendall $\tau_b$ \\ \midrule
  SOA$^\dagger$~\cite{hinz2020soa} & YOLO-V3 & 5.6 & 7.2 \\
  \midrule
  CLIP-S~\cite{Hessel2021} & CLIP (ViT-B/32) &  10.6 & 10.6 \\
  InfoNCE~\cite{oord2018infonce} & CLIP (ViT-B/32) & 10.9 & 10.9 \\
  CLIP-R-Precision~\cite{park2021benchmark} & CLIP (ViT-B/32) & 5.9 & 6.8 \\
  OFA-Captioning+CLIP-S~\cite{wang2022ofa,Hessel2021} & OFA-Large + CLIP (ViT-B/32) & 8.1 & 8.1 \\
  \midrule
  CLIP-S~\cite{Hessel2021} & CLIP (ViT-L/14) &  11.1 & 11.1 \\
  InfoNCE~\cite{oord2018infonce} & CLIP (ViT-L/14) & 11.1 & 11.1 \\
  CLIP-R-Precision~\cite{park2021benchmark} & CLIP (ViT-L/14) & 9.4 & 9.7 \\
  OFA-Captioning+CLIP-S~\cite{wang2022ofa,Hessel2021} & OFA-Large + CLIP (ViT-L/14) & 7.8 & 7.8 \\
  \midrule
  \pmi (ours) & CLIP (ViT-B/32) & \underline{11.9} & \underline{11.8} \\
  \pmi (ours) & CLIP (ViT-L/14) & \bf{12.4} & \bf{12.4} \\
  \bottomrule
  \end{tabular}
\end{table*}

%% file: tbl_reasoning.tex
\begin{table*}[t]
  \centering
  \small
  \caption{Visual reasoning accuracy using the foiled caption trick.} 
  \label{tbl:reasoning}
  \begin{tabular}{lcccccccc}
  \toprule
                   & Object   & Count & Color & Spatial & Object & Count & Color & Spatial \\  
  Metric           & \multicolumn{4}{c}{CLIP ViT-B/32} & \multicolumn{4}{c}{CLIP ViT-L/14} \\  \midrule
  CLIP-S~\cite{Hessel2021}                  & 0.318    & 0.026 & 0.068 & 0.025   & 0.585  & 0.157 & 0.209 & 0.169   \\
  CLIP-R-Precision~\cite{park2021benchmark} & 0.168    & 0.016 & 0.031 & 0.019   & 0.238  & 0.046 & 0.058 & 0.041   \\
  SOA~\cite{hinz2020soa}                    & 0.367    & 0.028 & 0.039 & 0.035   & 0.365  & 0.030 & 0.035 & 0.030   \\
  InfoNCE~\cite{oord2018infonce} & \U{0.416} & \U{0.042} & \U{0.094} & \U{0.049} & \U{0.675} & \U{0.230} & \U{0.284} & \U{0.243} \\  \midrule
  MID (ours) & {\bf 0.792} & {\bf 0.290} & {\bf 0.332} & {\bf 0.280} & {\bf 0.843} & {\bf 0.443} & {\bf 0.481} & {\bf 0.457} \\
  \bottomrule
  \end{tabular}
\end{table*}

%% file: tbl_flickr8k.tex
\begin{table*}[t!]
\begin{minipage}{.55\textwidth}
  \small
  \centering
  \caption{Flickr8K-Expert human judgment correlation.} 
  \label{tbl:flickr8k-expert}
  \begin{tabular}{lc}
  \toprule
  Method & Kendall $\tau_c$ \\ \midrule
  BLEU-1~\cite{papineni2002bleu} & 32.3 \\
  BLEU-4~\cite{papineni2002bleu} & 30.8 \\
  ROUGE-L~\cite{lin2004rouge} & 32.3 \\
  BERT-S (RoBERTa-F) & 39.2 \\
  METEOR~\cite{banerjee2005meteor} & 41.8 \\
  CIDEr~\cite{vedantam2015cider} & 43.9 \\
  SPICE~\cite{anderson2016spice} & 44.9 \\
  LEIC ($\tau_b$)~\cite{cui2018learning} & 46.6 \\
  BERT-S++~\cite{yi2020improving} & 46.7 \\
  TIGEr~\cite{jiang2019tiger} & 49.3 \\
  NUBIA~\cite{kane2020nubia} & 49.5 \\
  ViLBERTScore-F~\cite{lee2020vilbertscore} & 50.1 \\
  CLIP-S~\cite{Hessel2021} & 51.2 \\
  RefCLIP-S~\cite{Hessel2021} & \underline{53.0} \\
  \midrule
  \pmi (ours) & \bf{54.9} \\
  \bottomrule
  \end{tabular}
\end{minipage}
\hspace{1em}
\begin{minipage}{.4\textwidth}
  \small
  \centering
  \caption{Flickr8K-CF human judgment correlation.} 
  \label{tbl:flickr8k-cf}
  \begin{tabular}{lc}
  \toprule
  Method & Kendall $\tau_b$ \\ \midrule
  BLEU-4~\cite{papineni2002bleu}  & 16.9 \\
  CIDEr~\cite{vedantam2015cider}   & 24.6 \\
  METEOR~\cite{banerjee2005meteor}  & 22.2 \\
  ROUGE-L~\cite{lin2004rouge} & 19.9 \\
  SPICE~\cite{anderson2016spice}   & 24.4 \\
  BERT-S (RoBERTa-F) & 22.8 \\
  LEIC~\cite{cui2018learning}   & 29.5 \\
  CLIP-S~\cite{Hessel2021} & 34.4 \\
  RefCLIP-S~\cite{Hessel2021} & \underline{36.4} \\
  \midrule
  \pmi (ours) & \bf{37.3} \\
  \bottomrule
  \end{tabular}
\end{minipage}
\end{table*}

%% file: tbl_pascal_foil.tex
\begin{table*}[t!]
\begin{minipage}{.6\textwidth}
  \centering
  \small
  \caption{Pascal-50S accuracy. Please see the text for the definition of subsets, HC, HI, HM, and MM.} 
  \label{tbl:pascal}
  \begin{tabular}{lccccc}
  \toprule 
  Method & HC & HI & HM & MM & Mean \\
  \midrule
  length & 51.7 & 52.3 & 63.6 & 49.6 & 54.3 \\
  BLEU-4~\cite{papineni2002bleu} & 60.4 & 90.6 & 84.9 & 54.7 & 72.6 \\
  SPICE~\cite{anderson2016spice} & 63.6 & 96.3 & 86.7 & 68.3 & 78.7 \\
  METEOR~\cite{banerjee2005meteor} & 63.8 & 97.7 & 93.7 & 65.4 & 80.1 \\
  ROUGE-L~\cite{lin2004rouge} & 63.7 & 95.3 & 92.3 & 61.2 & 78.1 \\
  CIDEr~\cite{vedantam2015cider} & 65.1 & 98.1 & 90.5 & 64.8 & 79.6 \\
  BERT-S (RoBERTa-F) & \underline{65.4} & 96.2 & 93.3 & 61.4 & 79.1 \\
  TIGEr~\cite{jiang2019tiger} & 56.0 & \bf{99.8} & 92.8 & 74.2 & 80.7 \\
  ViLBERTScore-F~\cite{lee2020vilbertscore} & 49.9 & 99.6 & 93.1 & \underline{75.8} & 79.6 \\
  BERT-S++~\cite{yi2020improving} & \underline{65.4} & 98.1 & \underline{96.4} & 60.3 & 80.1 \\
  CLIP-S~\cite{Hessel2021} & 56.5 & 99.3 & \underline{96.4} & 70.4 & 80.7 \\
  RefCLIP-S~\cite{Hessel2021} & 64.5 & 99.6 & 95.4 & 72.8 & \underline{83.1} \\
  \midrule
  \pmi (ours) & \bf{67.0} & \underline{99.7} & \bf{97.4} & \bf{76.8} & \bf{85.2} \\
  \bottomrule
  \end{tabular}
\end{minipage}
\hspace{1.5em}
\begin{minipage}{.35\textwidth}
  \centering
  \small
  \caption{FOIL hallucination pairwise detection accuracy results. The methods utilize either one or four references.} 
  \label{tbl:foil}
  \begin{tabular}{lcc}
  \toprule
  Method & 1-ref & 4-ref \\
  \midrule
  length & 50.2 & 50.2 \\
  BLEU-4~\cite{papineni2002bleu} & 66.5 & 82.6 \\
  METEOR~\cite{banerjee2005meteor} & 78.8 & 85.4 \\
  ROUGE-L~\cite{lin2004rouge} & 71.7 & 79.3 \\
  CIDEr~\cite{vedantam2015cider} & 82.5 & 90.6 \\
  SPICE~\cite{anderson2016spice} & 75.5 & 86.1 \\
  BERT-S & 88.6 & \underline{92.1} \\
  CLIP-S~\cite{Hessel2021} & 87.2 & 87.2 \\
  RefCLIP-S~\cite{Hessel2021} & \bf{91.0} & \bf{92.6} \\
  \midrule
  \pmi (ours) & \underline{90.5} & 90.5 \\
  \bottomrule
  \end{tabular}
\end{minipage}
\end{table*}

%% file: 6_conclusions.tex
\label{sec:conclusions}
We, to our best knowledge, firstly argue that the negative cross-mutual information with multivariate Gaussian distributions can be used as a unified metric for multimodal generative models.
We provide the theoretical analyses of the proposed metric by out-of-distribution detecting by the squared Mahalanobis distances, bias and variance decomposition, and the relation to the Kullback-Leibler divergence, along with the empirical experiments.
We achieve the state-of-the-art performances on text-to-image generation and image captioning benchmarks, the generated and human Likert-scale judgment correlations, visual reasoning accuracy, Flickr8K-Expert, Flickr8K-CF, Pascal-50S, and FOIL hallucination detection.
We look forward to seeing the future works on the Gaussian cross-mutual information in multimodal representation learning based on this novel proposition.

%% file: appendix.tex
\section{Proofs}
\label{appendix:proofs}
\subsection{Proof of the mutual information with Gaussian distributions}
\label{appendix:mi-proof}
The mutual information of two Gaussian distributions is defined as:
\begin{align}
    I(\mX; \mY) = \frac{1}{2}\log\Big( \frac{\det(\Sigma_x) \det(\Sigma_y)}{\det(\Sigma_z)} \Big).
\end{align}

\begin{proof}
Let the mutual information be:
\begin{align}
    I(\mX; \mY) = \E_{p(\cond, \fake)} \Big[ \log \frac{p(\cond, \fake)}{p(\cond) p(\fake)} \Big].
\end{align}
Using the definition of multivariate Gaussian distribution as follows,
\begin{align}
    p(\vx)& = \frac{1}{\sqrt{(2\pi)^D \det(\Sigma)}} \exp \big[ -\frac{1}{2}(\vx - \mu)^\T \Sigma^{-1} (\vx - \mu) \big] 
\end{align}
we rearrange the equation to cancel out the constant terms. Then, the continuous mutual information is reduced to:
\begin{align}
    I(\mX; \mY) = \frac{1}{2}\log\Big( \frac{\det(\Sigma_x) \det(\Sigma_y)}{\det(\Sigma)} \Big) + \frac{1}{2} \E_{p(\x, \y)} \big[
        D_M^2(\x) + D_M^2(\y) - D_M^2(\z)
    \big] \label{eqn:e-pmi}
\end{align}
where $D_M^2$ denotes the squared Mahalanobis distance defined by $D_M^2(\x) = (\x - \mu_x)^\T \Sigma_\x^{-1} (\x - \mu_x)$, where $\mu_x$ and $\Sigma_\x$ are the mean and covariance of $\x$, and $\vz$ denotes $[\x; \y]$.

By the way, the expectation of the squared Mahalanobis distance is the dimension of samples, $D$.
\begin{align} \label{eqn:smd_is_d}
    \E_{p(\x)} D_M^2(\x) = \frac{1}{N} \tr\big(\mX^\intercal \Sigma_\x^{-1} \mX \big)
    = \frac{1}{N} \tr\big(\Sigma_\x^{-1} \mX \mX^\intercal \big)
    = \tr\big(\Sigma_\x^{-1} \Sigma_\x \big)
    = \tr( \I_D ) = D
\end{align}
where $\mX \in \R^{D \times N}$ is the samples, $\I_D \in \R^{D \times D}$ is the identity matrix. We use the cyclic property of trace where $\tr(ABC)=\tr(BCA)$.
Therefore, the second term reduces to zero as follows:
\begin{align}
    \frac{1}{2} \E_{p(\x, \y)} \big[D_M^2(\x) + D_M^2(\y) - D_M^2(\z)\big]
    = \frac{1}{2} (D + D - 2D) = 0.
\end{align}
We conclude the proof.
\end{proof}

By the way, the \textit{point-wise mutual information (PMI)} with Gaussian distributions can be derived from \eqn~\ref{eqn:e-pmi}:
\begin{align}
    \text{PMI}(\x; \y) = I(\mX; \mY) + \frac{1}{2} \big(
        D_M^2(\x) + D_M^2(\y) - D_M^2(\z)
    \big). 
\end{align}

\subsection{The bias-variance decomposition of the expectation of the squared Mahalanobis distance}
\label{appendix:bias-variance decompositon}

The expectation of PMI with respect to evaluating samples needs to calculate the expectation of three terms of the squared Mahalanobis distances (SMD) with respect to the evaluating sample $\hat{\x}$. With a notation of $\hat{\mX} \in \R^{D \times N}$ for N evaluation samples, we can decompose the expectation of SMD with two terms of bias and variance as follows:
\begin{align}
    \E_{\hat{\x}} \big[ D^2_M(\hat{\x}) \big] 
        &= \frac{1}{N}\tr\big( (\hat{\mX} - \mu_\x \One^\intercal)^\intercal \Sigma_\x^{-1} (\hat{\mX} - \mu_\x \One^\intercal) \big) \\
        &= \frac{1}{N} \tr\big(\Sigma_\x^{-1} (\hat{\mX} - \mu_\x \One^\intercal) (\hat{\mX} - \mu_\x \One^\intercal)^\intercal \big) \\
        &= \frac{1}{N} \tr\Big(\Sigma_\x^{-1} \big(\hat{\mX}\hat{\mX}^\intercal - N \mu_{\hat{\x}}\mu_{\hat{\x}}^\intercal + N( \mu_{\hat{\x}} - \mu_{\x})(\mu_{\hat{\x}} - \mu_{\x})^\intercal \big) \Big) \\
        &= \tr\Big(\Sigma_\x^{-1} \big(\Sigma_{\hat{\x}} + ( \mu_{\hat{\x}} - \mu_{\x})(\mu_{\hat{\x}} - \mu_{\x})^\intercal \big) \Big) \\
        &= (\mu_{\hat{\x}} - \mu_{\x})^\intercal \Sigma_\x^{-1}( \mu_{\hat{\x}} - \mu_{\x}) + \tr(\Sigma_\x^{-1}\Sigma_{\hat{\x}}) \\
        &= (\mu_{\hat{\x}} - \mu_{\x})^\intercal \Sigma_\x^{-1}( \mu_{\hat{\x}} - \mu_{\x}) + \tr(\Sigma_\x^{-1}\Sigma_{\hat{\x}})
        - \tr(\Sigma_\x^{-1}\Sigma_{\x}) + \tr(\Sigma_\x^{-1}\Sigma_{\x}) \\
        &= (\mu_{\hat{\x}} - \mu_{\x})^\intercal \Sigma_\x^{-1}( \mu_{\hat{\x}} - \mu_{\x}) + \tr\big(\Sigma_\x^{-1}(\Sigma_{\hat{\x}} - \Sigma_{\x})\big) + D \label{eqn:smd_kl}
\end{align}
where $\One \in \R^{N}$ is a vector of ones. Remind that the expectation of SMD is $D$ when the evaluating samples $\hat{\x}$ are following the distribution of $\x$ in \eqn~\ref{eqn:smd_is_d}.
However, the above equation shows that if the mean or covariance of $\hat{\x}$ deviates from $\x$, the result may be smaller or larger than D.

\subsection{Relation to Kullback–Leibler divergence}
\label{appendix:kl-divergence}
The proposed method \pmi is related to Kullback-Leibler divergence (or relative entropy).
Let $\N_0(\mu_0, \Sigma_0)$ and $\N_1(\mu_1, \Sigma_1)$ are two multivariate normal distributions having the same dimension of $D$, then the Kullback-Leibler divergence between the distributions is as follows~\cite{duchi2007kld}:
\begin{align}
    \displaystyle D_{\text{KL}}\left({\mathcal {N}}_{0}\parallel {\mathcal {N}}_{1}\right)={\frac {1}{2}}\left(\operatorname {tr} \left(\Sigma _{1}^{-1}(\Sigma _{0}-\Sigma _{1})\right)+\left(\mu _{1}-\mu _{0}\right)^{\mathsf {T}}\Sigma _{1}^{-1}\left(\mu _{1}-\mu _{0}\right) + \log \left({\frac{\det \Sigma _{1}}{\det \Sigma _{0}}}\right)\right). \nonumber
\end{align}
Using the above equation and \eqn~\ref{eqn:smd_kl}, we rearrange \eqn~\ref{eqn:e-pmi} as follows:
\begin{align}
    \E_{(\hat{\x}, \hat{\y}) \sim \mathcal{D}} \text{PMI}(\hat{\x}; \hat{\y}) 
        = & I(\mX; \mY)
        + D_{\text{KL}}(p(\hat{\x}) \parallel p(\x) ) + D_{\text{KL}}(p(\hat{\y}) \parallel p(\y) ) - D_{\text{KL}}(p(\hat{\z}) \parallel p(\z) ) \nonumber \\
          & - \frac{1}{2} \Big( 
            \log \Big( \frac{\det \Sigma_{\x}}{\det \Sigma_{\hat{\x}}} \Big)
            + \log \Big( \frac{\det \Sigma_{\y}}{\det \Sigma_{\hat{\y}}} \Big)
            - \log \Big( \frac{\det \Sigma_{\z}}{\det \Sigma_{\hat{\z}}} \Big) \Big) \\
        = & I(\mX; \mY)
        + D_{\text{KL}}(p(\hat{\x}) \parallel p(\x) ) + D_{\text{KL}}(p(\hat{\y}) \parallel p(\y) ) - D_{\text{KL}}(p(\hat{\z}) \parallel p(\z) ) \nonumber \\
          & - \frac{1}{2} \log \Big( \frac{\det\Sigma_{\x}\det\Sigma_{\y}}{\det \Sigma_{\z}} \Big)
          + \frac{1}{2} \log \Big( \frac{\det\Sigma_{\hat{\x}}\det\Sigma_{\hat{\y}}}{\det \Sigma_{\hat{\z}}} \Big) \\
        = & I(\hat{\mX}; \hat{\mY}) + D_{\text{KL}}(p(\hat{\x}) \parallel p(\x) ) - D_{\text{KL}}(p(\hat{\z}) \parallel p(\z) )
        \label{eqn:mid_kl}
\end{align}
where $D_{\text{KL}}(p(\hat{\y}) \parallel p(\y) ) = 0$ since $\hat{\y}$ and $\y$ are the same condition evaluating generations.


\section{Generated Likert-scale judgment correlation using VQ-Diffusion}

\begin{figure}
  \begin{center}
    \includegraphics[width=\textwidth]{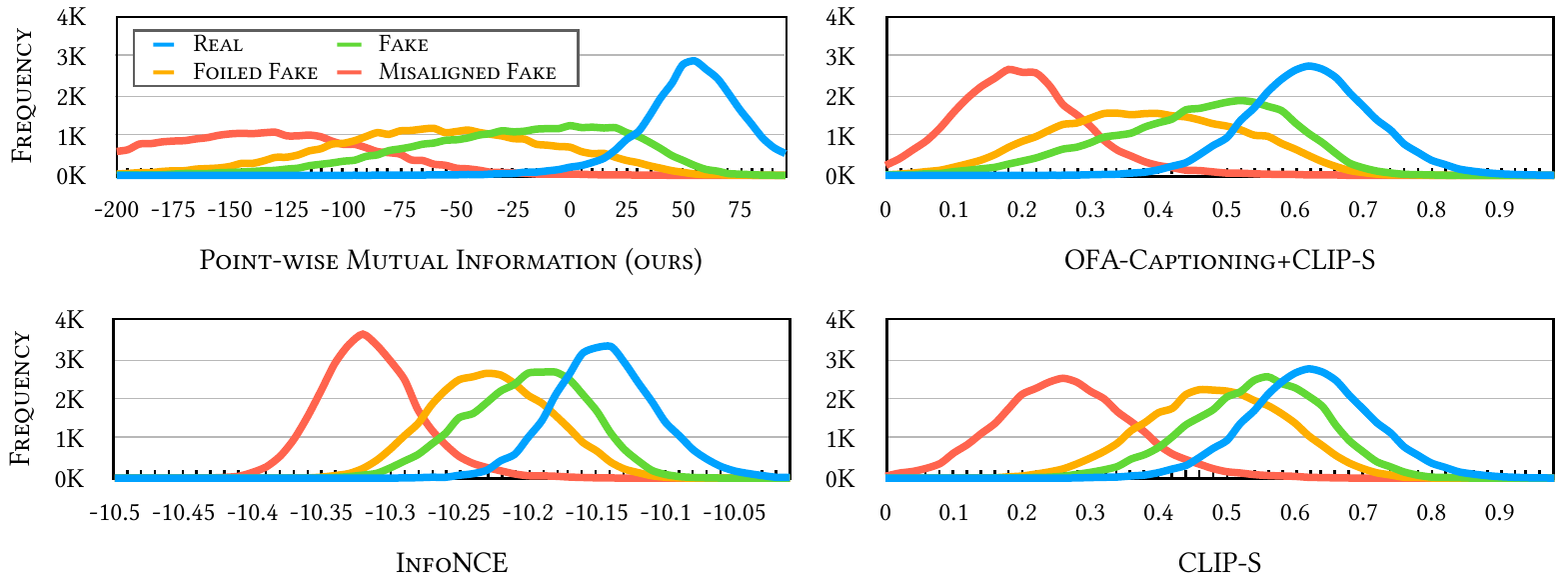}
  \end{center}
  \caption{The histogram shows the frequencies of the competing methods using the CLIP of ViT-L/14 for the four-scale judgments from the COCO dataset. The real images and the original captions ({\bf Real}), the fake images generated from the VQ-Diffusion~\cite{gu2021vqdiffusion} using the original caption ({\bf Fake}), the fake images generated from the VQ-Diffusion but using the foiled caption deliberately swapping some objects to the other words ({\bf Foiled Fake}) (notice that the foiled fake images are still matching with the original captions to assess), and randomly matching the foiled fake to the original captions by shuffling ({\bf Misaligned Fake}). For the detail of methods, please see the text in \sect~\ref{sec:experiment}.
  }
  \label{fig:likert_vqdiffusion}
\end{figure}

\tbl~\ref{tbl:likert-vqdiffusion} and \fig~\ref{fig:likert_vqdiffusion} show the results from the (foiled) fake images using VQ-Diffusion~\cite{gu2021vqdiffusion}.
While the proposed MID outperforms the competing methods, the portion of fake images that get higher scores than real images is decreased in InfoNCE and CLIP-S.
This observation may attribute to the under-performance of VQ-Diffusion than LAFTIE or the side effect of the contrastive loss used in LAFITE. Remind that our method shows the consistency toward different models among the comparative metrics.

\input{tbl_likert_vqdiffusion}

\section{The details on visual reasoning accuracy}
\label{appendix:reasoning}
We describe the detail of visual reasoning accuracy in \tbl~\ref{tbl:reasoning}. For the object task, we use randomly sampled 30K captions from the FOIL dataset~\cite{shekhar2017foil}. For the count task, we use a set of tokens "0", "1", "2", "3", "4", "one", "two", "three", and "four". For the color task, we use the sixteen basic color keywords~\footnote{\url{https://www.w3.org/TR/css-color-3/\#html4}}. For the spatial relationship task, we use "above", "below", "left", "right", "front", and "back". The number of samples are 30K, 1.3K, 4.6K, 1.5K for the object, count, color, spatial relationship tasks, respectively. The LAFITE~\cite{zhou2021lafite} generates the (foiled) fake images.


\begin{figure}[ht!]
  \begin{center}
    \includegraphics[width=.8\textwidth]{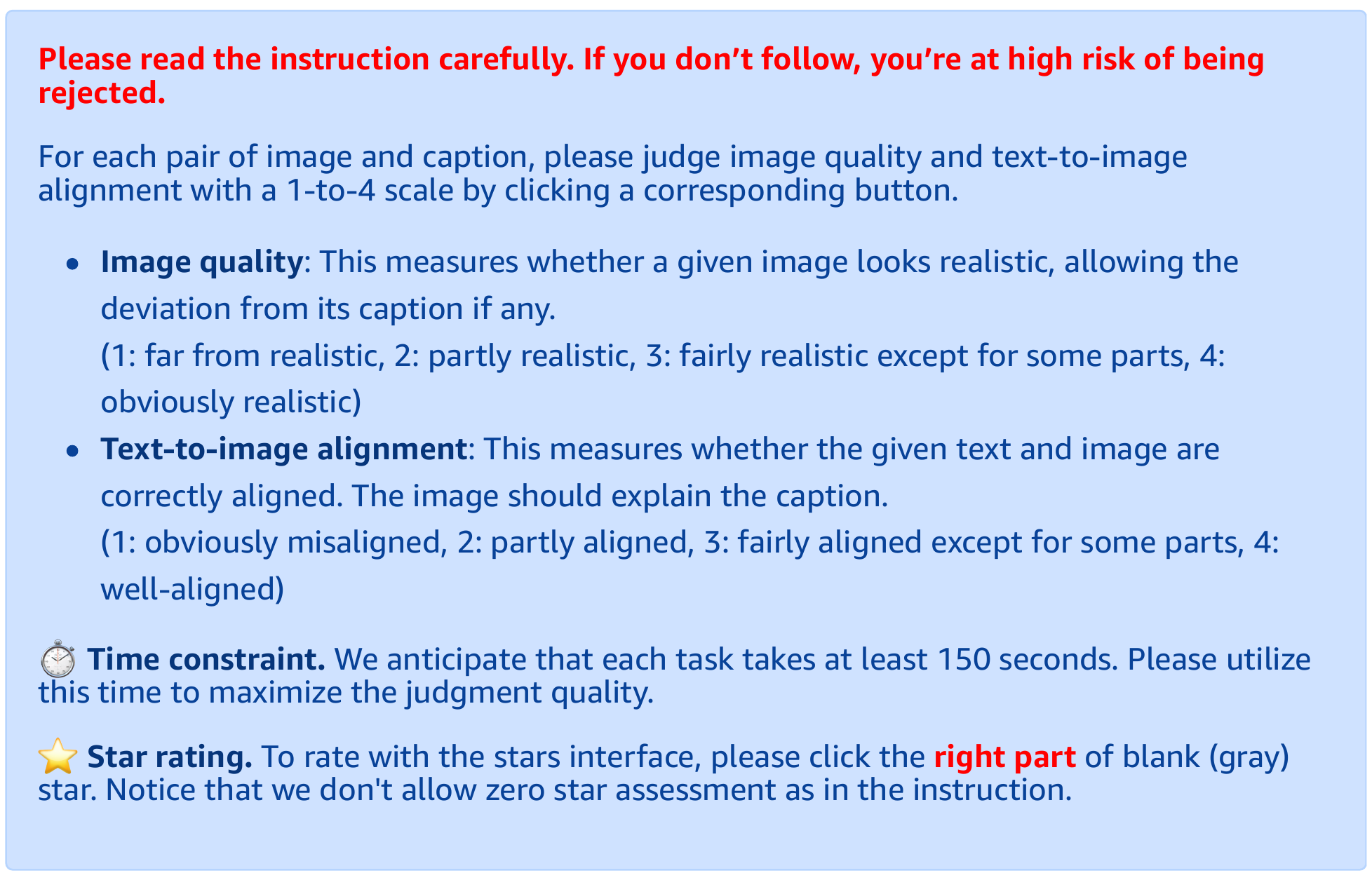}
  \end{center}
  \caption{The instruction appeared on top of the AMT task interface.}
  \label{fig:hit_instruction}
\end{figure}
\begin{figure}[ht!]
  \begin{center}
    \includegraphics[width=.7\textwidth]{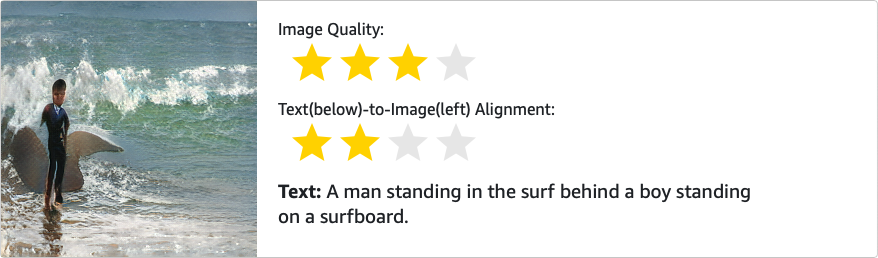}
  \end{center}
  \caption{The one of ten samples in an assignment.}
  \label{fig:task_one_of_ten}
\end{figure}

\section{Human judgment collection from the Amazon Mechanical Turk}
\label{appendix:amt}
In \sect~\ref{sec:amt}, the human Likert-scale judgments on the fine-grained text-image alignment are used to evaluate the metrics and reported the rank correlations in \tbl~\ref{tbl:likert-amt}.
The following paragraphs describe the data collection procedure using the Amazon Mechanical Turk (AMT).

\paragraph{Data.}
We randomly sample 1K validation captions from the COCO dataset~\cite{chen2015coco} to generate fake images using the LAFITE~\cite{zhou2021lafite} and VQ-Diffusion~\cite{gu2021vqdiffusion}, which makes 2K fake images in total.
Each \textbf{task} consists of 10 randomly-sampled fake images and the corresponding captions. A worker from the AMT is carefully instructed to annotate the visual quality of each fake image and the text-image alignment between the fake image and its caption. The visual quality annotation is designed as a preliminary task to reduce the quality bias in the text-image alignment assessment. There are five \textbf{assignments} per \textbf{task} to collect five annotations to decide a final judgment score for the image. We used the median of five annotations, which makes 2K evaluating samples on 1K captions. For the reference samples, we use randomly-sampled 30K captions and the corresponding images from the validation split of the COCO dataset as in the generated Likert-scale judgment experiments.

\paragraph{Interface.} Each \textbf{task} presents with the instruction on top of the AMT task interface as in \fig~\ref{fig:hit_instruction}. We adopt a one-to-four Likert scale to assess the image quality and text-to-image alignment. There are descriptions of how to interpret each level of score.
We encourage the workers to spend at least 150 seconds for 20 assessments (10 images, two assessments for each image).
We observed a few mild violations of this time constraint in a pilot deployment, so we systemically forced the workers to spend at least 150 seconds before the submission. \fig~\ref{fig:task_one_of_ten} shows one of ten samples in a \textbf{assignment}.

\begin{figure}[ht!]
  \begin{center}
    \includegraphics[width=\textwidth]{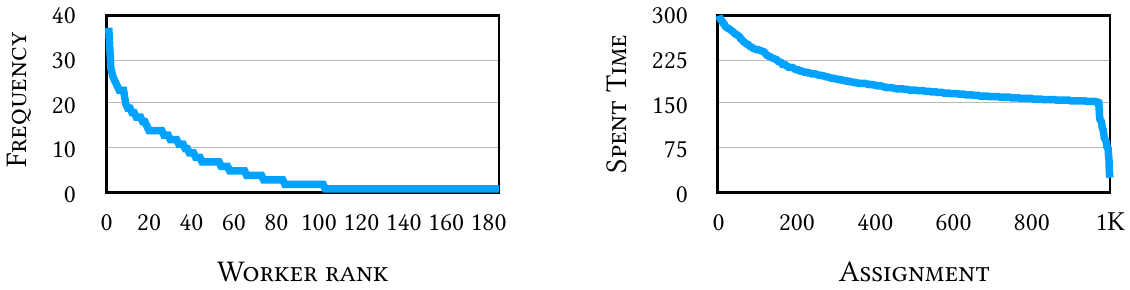}
  \end{center}
  \caption{\textbf{Left.} The number of assignments per worker. \textbf{Right.} Time spent per assignment.}
  \label{fig:workers}
\end{figure}

\paragraph{Cost.} We collected 1K \textbf{assignments} for 200 \textbf{tasks}. Considering reasonable earn per hour, we paid \$0.21 for each assignment, \$210 in total.

\paragraph{Statistics.} \fig~\ref{fig:workers} shows the number of assignments per worker (left) and the time spent per assignment. A worker did at most 37 assignments and the assignments are done within five minutes.
After taking median, the mean and standard deviation of the quality judgments are 2.43 and 0.61, respectively, and these of the alignment judgments are 2.62 and 0.64, respectively.

\begin{figure*}[ht!]
  \begin{center}
    \includegraphics[width=\textwidth]{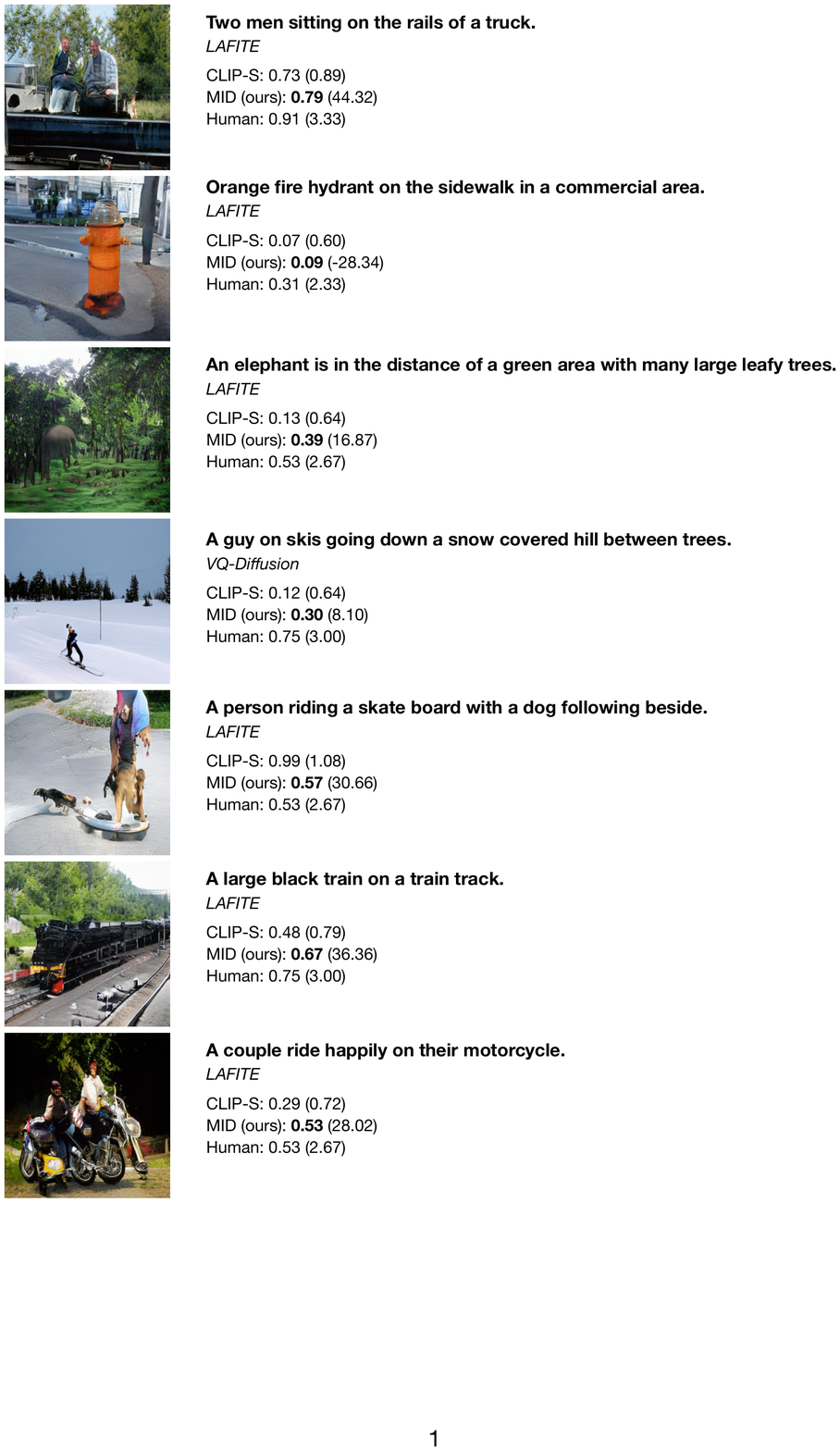}
  \end{center}
  \caption{The visualization of generated images and their evaluation results. From the top, a caption for generation, a generative model type, a normalized rank and a raw score in parentheses for each metric. The normalized rank denotes the evaluated score rank divided by the number of samples (zero-to-one scale) for a fair comparison. \textbf{bold} indicates the closer to human judgments.}
  \label{fig:amt_vis}
\end{figure*}

\paragraph{Visualization} \fig~\ref{fig:amt_vis} shows some examples comparing with human judgment scores. Note that we used the mean of three median annotations from workers for this visualization. 

\section{Performance results of text-to-image generative models}
\label{appendix:gans}
\input{tbl_gans}
\tbl~\ref{tbl:gans} illustrates the performance of text-to-image generative models including the proposed MID scores. The reports of FID, SOA-C, and SOA-I are from \citet{hinz2020soa} while missing scores are from the corresponding cites. They ``randomly sampled three times 30,000 images from the training set and compared them to the statistics of the validation set.''
We found that LAFITE uses all 82,612 training images to get the statistics, so we also report FID (ours) for a fair comparison. We sampled 30,000 fake images from the validation captions.
This different sampling strategy may attribute to the difference of the real images' upper bound (6.09 vs. 2.73).
For the proposed MID, we randomly sampled 30,000 images and the captions for each image from the validation set as reference samples ($\mX, \mY)$.
We trained the LAFITE model from scratch using the official code~\footnote{\url{https://github.com/drboog/Lafite}} with the same hyper-parameters and 1.5 times training longer to achieve a slightly better FID than the publicly released model.
Interestingly, the filtered GLIDE underperforms with the worst FID; however, it outperforms some of the other models with MID. It may show that the data filtering severely affects FID while relatively retaining the performance of text-image alignment captured by MID.


\section{Performance comparison on the comp-t2i dataset}
\label{appendix:comp-t2i}

\tbl~\ref{tbl:compt2i} demonstrates the human judgement correlation for the comp-t2i dataset~\cite{park2021benchmark}.
We compute Pearson correlation coefficient (PCC), Spearman correlation coefficient (SCC), binary decision consensus accuracy (Acc.), and Kendall's $\tau$ coefficients between metric scores and human judgment scores.
The human judgment scores are pre-processed using the ratio of $n/5$ where $n$ is the number of votes.
Pearson correlation coefficient is a statistic that measures the linear correlation, while Spearman correlation coefficient evaluates a monotonic relationship rather than the raw values.
Following~\cite{park2021benchmark}, the accuracy is measured for binary decision consensus for the seen and swapped captions.
Additionally, we report two variants of the Kendall's $\tau$ having a better confidence interval for more precise comparison.

We clarify that even with the released asset, we cannot reproduce the reported scores for the CLIP-R-Precision of some splits. So, we report our reproduction along with their reported scores. Following the released code, we exploit available bounding boxes for the image pre-processing and sample the negative examples from a randomly-chosen different class for the CLIP-R-Precision. Notice that CLIP-R-Precision has a moderate variance since the sampled negative examples impact the score while risking some degree of false negative describing narrow domains of birds and flowers.

We conducted the experiment using both the pre-trained CLIP and the CLIP model further fine-tuned on the corresponding dataset. 
Note that CLIP model is based on ResNet 101.
Since CLIP-R-Precision score is determined by a set of caption candidates, we repeat 10 times to construct the negative examples and report standard deviation.
To measure the proposed MID, the training set for fine-tuning  CLIP is used as the reference set ($\mX, \mY$) in \eqn~\ref{eqn:mid_kl} (\Appendix), while the scored image-caption pairs are the evaluation samples ($\hat{\mX}, \hat{\mY}$).
In most cases, regardless of whether it is fine-tuned or not, the proposed MID outperforms CLIP-S and CLIP-R-Precision.
These results suggest that our MID has better generalization capabilities.

\input{tbl_compt2i}


\section{Utilizing the references like RefCLIP-S for FOIL hallucination detection}
\label{appendix:extra-foil}
When we similarly utilize the term of $\max(\max_{\vr \in \mR} \cos(\vr, \vy), 0)$ in \eqn~\ref{eqn:refclip-s} for \pmi, it results in 92.4 and 93.7, outperforming RefCLIP-S. Since the scoring scale of \pmi is different from cosine similarity, we use the arithmetic mean instead of the harmonic mean and a different weight $\alpha$ of 3e2 considering the difference between their standard deviations as follows:
\begin{align}
    \text{RefMID}(\vx, \vy, \mR) = \frac{1}{2} \big(\text{MID}(\vx, \vy), \alpha \max(\max_{\vr \in \mR} \cos(\vr, \vy), 0)\big)
\end{align}
where MID is parameterized by the referencing image and caption statistics. Notice that RefCLIP-S also has the hyper-parameter of $2.5$ in the CLIP-S to balance with the cosine similarity term.

\begin{figure}[ht!]
  \begin{center}
    \includegraphics[width=\textwidth]{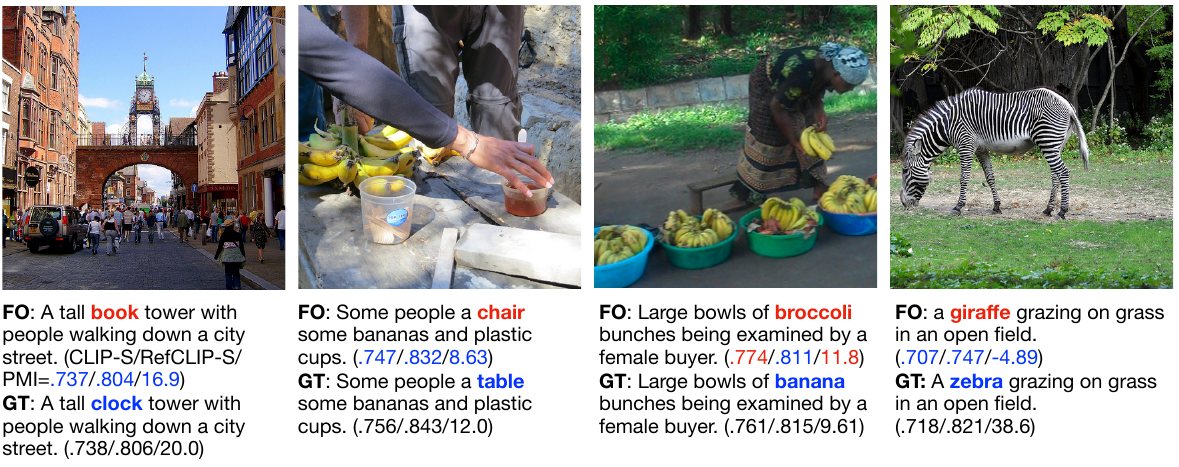}
  \end{center}
  \caption{The visualization of the FOIL hallucination detection. {\bf FO} and {\bf GT} denote FOIL and ground-truth captions, respectively. The foiled word is highlighted by red and its counterpart is by blue. For each caption, we report the scores of CLIP-S~\cite{Hessel2021}, RefCLIP-S~\cite{Hessel2021}, and \pmi (ours). First two columns show the corrected examples, while the third column shows an example that CLIP-S and \pmi are failed to detect. The smaller score is better for the FOIL captions. As pointed out in \sect~\ref{sec:evaluation_on_image_captioning}, RefCLIP-S directly exploits the reference captions where the counterpart of the foiled word is appeared. One of references of the third example was that ``a woman is picking {\it bananas} from a basket.'' The fourth example shows that \pmi can be negative for unlikely samples since it is based on the definition of differential entropy.}
  \label{fig:vis_foil}
\end{figure}

%% file: tbl_likert_vqdiffusion.tex
\begin{table*}[h!]
  \centering
  \small
  \caption{Generated Likert-scale judgment correlation using VQ-Diffusion. $\dagger$ uses the i.i.d. samples having at least one detected object, which was 88.1\%, to calculate the SOA accuracy per image.} 
  \label{tbl:likert-vqdiffusion}
  \begin{tabular}{lccc}
  \toprule
  Method & Backbone & Kendall $\tau_c$ & Kendall $\tau_b$ \\ \midrule
  SOA$^\dagger$~\cite{hinz2020soa} & YOLO-V3 & 37.0 & 38.4 \\
  \midrule
  CLIP-S~\cite{Hessel2021} & CLIP (ViT-B/32) &  70.3 & 60.9 \\
  InfoNCE~\cite{oord2018infonce} & CLIP (ViT-B/32) & 74.2 & 64.3 \\
  CLIP-R-Precision~\cite{park2021benchmark} & CLIP (ViT-B/32) & 66.5 & 54.5 \\
  OFA-Captioning+CLIP-S~\cite{wang2022ofa,Hessel2021} & OFA-Large + CLIP (ViT-B/32) & 73.8 & 63.9 \\
  \midrule
  CLIP-S~\cite{Hessel2021} & CLIP (ViT-L/14) &  70.9 & 61.4 \\
  InfoNCE~\cite{oord2018infonce} & CLIP (ViT-L/14) & 78.0 & 67.6 \\
  CLIP-R-Precision~\cite{park2021benchmark} & CLIP (ViT-L/14) & 68.5 & 56.5 \\
  OFA-Captioning+CLIP-S~\cite{wang2022ofa,Hessel2021} & OFA-Large + CLIP (ViT-L/14) & 74.2 & 64.3 \\
  \midrule
  \pmi (ours) & CLIP (ViT-B/32) & \underline{79.8} & \underline{69.1} \\
  \pmi (ours) & CLIP (ViT-L/14) & \bf{82.0} & \bf{71.1} \\
  \bottomrule
  \end{tabular}
\end{table*}

%% file: tbl_gans.tex
\begin{table*}[h]
  \centering
  \small
  \caption{The performance reports of text-to-image generative models. $^\dagger$The publicly-released model using the filtered subset of training dataset (\url{https://github.com/openai/glide-text2im}).
  For the backbone, I, Y, C stands for Inception-V3~\cite{szegedy2016inceptionv3}, YOLO-V3~\cite{redmon2018yolov3}, and CLIP~\cite{Radford2021}.
  }
  \label{tbl:gans}
  \begin{tabular}{lccccccc}
  \toprule
Metric & FID $\downarrow$ & SOA-C $\uparrow$ & SOA-I $\uparrow$ & FID (ours) $\downarrow$ & MID $\uparrow$ & MID $\uparrow$ \\
Backbone & I & Y & Y & I & C (ViT-B/32) & C (ViT-L/14) \\
\midrule
GLIDE~\cite{nichol2021glide}$^{\dagger}$ &
-            & -     & -     & 32.08\C{.05} &{\mz}1.00\C{.16}  &{\mz}1.03\C{.06} \\
\midrule
AttnGAN~\cite{xu2018attngan} &
33.10\C{.11} & 25.88 & 39.01 & 29.15\C{.06} &{\nz}-8.90\C{.18} & -65.20\C{.92} \\ 
DM-GAN~\cite{zhu2019dmgan} &
27.34\C{.11} & 33.44 & 48.03 & 22.90\C{.06} &{\mz}3.51\C{.20} & -44.66\C{.71} \\
OP-GAN~\cite{hinz2020soa} &
24.70\C{.09} & 35.85 & 50.47 & 22.14\C{.01} &{\nz}-1.32\C{.10} & -50.34\C{.84} \\
DF-GAN~\cite{tao2020dfgan} &
21.42\X{.00} & -     & -     & 31.75\C{.06} & -15.21\C{.12} & -58.75\C{.16} \\ 
VQ-Diffusion~\cite{gu2021vqdiffusion} &
13.86\X{.00} & -     & -     & 13.13\C{.05} &{\mz}5.77\C{.11} & -19.63\C{.28} \\
LAFITE~\cite{zhou2021lafite} & 
{\nz}\textbf{8.12}\X{.00} & \textbf{61.09} & \textbf{74.78} & {\nz}\textbf{8.03\C{.01}} & {\mn}\textbf{35.17\C{.20}} & {\mz}\textbf{6.26\C{.69}} \\
\midrule
Real & 
6.09\C{.05} & 74.97      & 80.84      & {\nz}2.73\C{.15} & 41.63\C{.06} & {\mn}57.44\C{.06} \\
  \bottomrule
  \end{tabular}
\end{table*}

%% file: tbl_compt2i.tex
\begin{table*}[h]
  \centering
  \small
  \caption{The comp-t2i human judgment correlation~\cite{park2021benchmark}. `FT' denotes whether the pre-trained CLIP model is further fine-tuned on the corresponding dataset which is available at \url{{https://github.com/Seth-Park/comp-t2i-dataset}}, and $\dagger$ denotes our reproduction.
  The best result and the second best result are boldfaced and underlined, respectively.}
  \label{tbl:compt2i}
  \begin{tabular}{llcccccc}
  \toprule
  & Metric & FT & PCC & SCC & Acc. & $\tau_c$ & $\tau_b$ \\
  \midrule
  \multirow{8}{*}{\rotatebox{90}{C-CUB Color}} & Human~\cite{park2021benchmark} & - & 0.5949\X{.000} & 0.5890\X{.000} & 81.7\X{.00} & - & -\\
  & DAMSM~\cite{xu2018attngan,park2021benchmark} & - & 0.0503\X{.000} & 0.1224\X{.000} & 54.9\X{.00} & - & - \\
  \cmidrule(lr){2-8}
    & CLIP-S~\cite{Hessel2021} & \xmark 
    & \underline{0.1919}\X{.000} & \underline{0.1865}\X{.000} & \underline{67.0}\X{.00} & \underline{14.78}\X{.000} & \underline{13.59}\X{.000}\\
    & CLIP-R-Precision~\cite{park2021benchmark}$\dagger$ & \xmark 
    & 0.1410\C{.013} & 0.1426\C{.014} & 53.3\C{1.3} & 13.02\C{1.25} & 12.62\C{1.27}\\
    & MID (ours) & \xmark
    & \textbf{0.2863}\X{.000} & \textbf{0.3499}\X{.000} & \textbf{68.5}\X{.00} & \textbf{27.68}\X{.000} & \textbf{25.45}\X{.000}\\
  \cmidrule(lr){2-8}
    & CLIP-S~\cite{Hessel2021} & \cmark 
    & \textbf{0.3711}\X{.000} & \underline{0.3428}\X{.000} & \textbf{77.5}\X{.00} & 27.24\X{.000} & \underline{25.05}\X{.000} \\
    & CLIP-R-Precision~\cite{park2021benchmark} & \cmark 
    & 0.0752\X{.000} & 0.1263\X{.000} & 56.4\X{.00} & - & -\\
    & CLIP-R-Precision~\cite{park2021benchmark}$\dagger$ & \cmark 
    & 0.2538\C{.014} & 0.2474\C{.014} & 62.1\C{1.4} & \underline{28.04}\C{1.52} & 21.90\C{1.19}\\
    & MID (ours) & \cmark  
    & \underline{0.3558}\X{.000} & \textbf{0.3990}\X{.000} & \underline{76.5}\X{.00} & \textbf{31.80}\X{.000} & \textbf{29.24}\X{.000}\\
    
  \midrule
  \multirow{8}{*}{\rotatebox{90}{C-CUB Shape}} & Human~\cite{park2021benchmark} & - & 0.3949\X{.000} & 0.4007\X{.000} & 70.0\X{.00} & - & - \\
  & DAMSM~\cite{xu2018attngan,park2021benchmark} & - & 0.1229\X{.000} & 0.0170\X{.000} & 52.7\X{.00} & - & - \\
  \cmidrule(lr){2-8}
    & CLIP-S~\cite{Hessel2021} & \xmark 
    & \underline{0.0287}\X{.000} & \underline{0.0315}\X{.000} & \textbf{61.0}\X{.00} & \underline{2.37}\X{.000} & \underline{2.17}\X{.000} \\
    & CLIP-R-Precision~\cite{park2021benchmark}$\dagger$ & \xmark 
    & 0.0071\C{.012} & 0.0078\C{.012} & 43.9\C{0.6} & {\nz}0.7\C{1.07} & 0.69\C{1.05}\\
    & MID (ours) & \xmark
    & \textbf{0.1113}\X{.000} & \textbf{0.1079}\X{.000} & \underline{55.0}\X{.00} & \textbf{8.56}\X{.000} & \textbf{7.85}\X{.000} \\
  \cmidrule(lr){2-8}
    & CLIP-S~\cite{Hessel2021} & \cmark 
    & 0.0577\X{.000} & 0.0593\X{.000} & \underline{56.5}\X{.00} & {\nz}4.64\X{.000} & 4.26\X{.000} \\
    & CLIP-R-Precision~\cite{park2021benchmark} & \cmark
    & 0.0878\X{.000} & 0.0806\X{.000} & 52.2\X{.00} & - & -\\
    & CLIP-R-Precision~\cite{park2021benchmark}$\dagger$ & \cmark 
    & \underline{0.1096}\C{.022} & \underline{0.1063}\C{.022} & 46.3\C{1.0} & \textbf{11.97\C{2.40}} & \textbf{9.40\C{1.90}} \\
    & MID (ours) & \cmark  
    & \textbf{0.1118}\X{.000} & \textbf{0.1280}\X{.000} & \textbf{60.0}\X{.00} & {\nz}\underline{9.96}\X{.000} & \underline{9.14}\X{.000} \\
    
  \midrule
  \multirow{8}{*}{\rotatebox{90}{C-Flower Color}} & Human~\cite{park2021benchmark} & -
  & 0.5891\X{.000} & 0.5870\X{.000} & 80.5\X{.00} & - & - \\
  & DAMSM~\cite{xu2018attngan,park2021benchmark} & - & -0.0457\X{.0000} & 0.0456\X{.000} & 52.3\X{.00} & - & - \\
  \cmidrule(lr){2-8}
    & CLIP-S~\cite{Hessel2021} & \xmark 
    & \underline{0.1802}\X{.000} & \underline{0.1823}\X{.000} & \textbf{70.0}\X{.00} & \underline{14.29}\X{.000} & \underline{13.13}\X{.000}\\
    & CLIP-R-Precision~\cite{park2021benchmark}$\dagger$ & \xmark 
    & 0.0939\C{.011} & 0.0943\C{.012} & 49.5\C{0.5} & {\nz}8.95\C{1.13} & {\nz}8.34\C{1.04}\\
    & MID (ours) & \xmark 
    & \textbf{0.2872}\X{.000} & \textbf{0.3433}\X{.000} & \underline{69.0}\X{.00} & \textbf{26.95}\X{.000} & \textbf{24.76}\X{.000} \\
  \cmidrule(lr){2-8}
    & CLIP-S~\cite{Hessel2021} & \cmark 
    & \textbf{0.4486}\X{.000} & \textbf{0.4321}\X{.000} & \textbf{81.0}\X{.00} & \textbf{34.85}\X{.000} & \textbf{32.03}\X{.000} \\
    & CLIP-R-Precision~\cite{park2021benchmark} & \cmark 
    & 0.2818\X{.000} & 0.3151\X{.000} & 65.3\X{.00} & - & -\\
    & CLIP-R-Precision~\cite{park2021benchmark}$\dagger$ & \cmark 
    & 0.2892\C{.014} & 0.2864\C{.014} & 58.8\C{1.2} & 31.04\C{1.51} & 25.33\C{1.25}\\
    & MID (ours) & \cmark 
    & \underline{0.4156}\X{.000} & \underline{0.4289}\X{.000} & \underline{74.0}\X{.00} & \underline{34.48}\X{.000} & \underline{31.68}\X{.000} \\  

    \midrule
    \multirow{8}{*}{\rotatebox{90}{C-Flower Shape}} & Human~\cite{park2021benchmark} & -
    & 0.3721\X{.000} & 0.3698\X{.000} & 68.8\X{.00} & - & - \\
    & DAMSM~\cite{xu2018attngan,park2021benchmark} & - & -0.0435\X{.0000} & 0.0876\X{.000} & 48.0\X{.00} & - & - \\
    \cmidrule(lr){2-8}
    & CLIP-S~\cite{Hessel2021} & \xmark 
    & 0.0330\X{.000} & 0.0300\X{.000} & \underline{54.0}\X{.00} & 2.31\X{.000} & 2.12\X{.000} \\
    & CLIP-R-Precision~\cite{park2021benchmark}$\dagger$ & \xmark 
    & \underline{0.0613}\C{.016} & \underline{0.0604}\C{.016} & 46.4\C{1.8} & \underline{5.99}\C{1.65} & \underline{5.34}\C{1.45} \\
    & MID (ours) & \xmark 
    & \textbf{0.0893}\X{.000} & \textbf{0.0994}\X{.000} & \textbf{59.0}\X{.00} & \textbf{7.93}\X{.000} & \textbf{7.28}\X{.000} \\
    \cmidrule(lr){2-8}
    & CLIP-S~\cite{Hessel2021} & \cmark 
    & 0.0875\X{.000} & \underline{0.0809}\X{.000} & \underline{57.0}\X{.00} & \underline{6.40}\X{.000} & \underline{5.87}\X{.000} \\
    & CLIP-R-Precision~\cite{park2021benchmark} & \cmark 
    & \underline{0.0920}\X{.000} & 0.0702\X{.000} & 52.0\X{.00} & - & -\\
    & CLIP-R-Precision~\cite{park2021benchmark}$\dagger$ & \cmark 
     & 0.0467\C{.017} & 0.0473\C{.017} & 46.2\C{1.0} & 5.38\C{1.92} & 4.18\C{1.49} \\
    & MID (ours) & \cmark 
    & \textbf{0.1056}\X{.000} & \textbf{0.1212}\X{.000} & \textbf{63.5}\X{.00} & \textbf{9.66}\X{.000} & \textbf{8.87}\X{.000} \\  

  \bottomrule
  \end{tabular}
\end{table*}